\documentclass[letterpaper]{article} 
\usepackage{aaai2026}  
\usepackage{times}  
\usepackage{helvet}  
\usepackage{courier}  
\usepackage[hyphens]{url}  
\usepackage{graphicx} 
\usepackage{natbib}  
\usepackage{caption} 
\usepackage{algorithm}
\usepackage{algorithmic}
\usepackage{times}
\usepackage{helvet}
\usepackage{courier}
\usepackage{xcolor}

\usepackage{newfloat}
\usepackage{listings}
\DeclareCaptionStyle{ruled}{labelfont=normalfont,labelsep=colon,strut=off} 
\floatstyle{ruled}
\newfloat{listing}{tb}{lst}{}
\floatname{listing}{Listing}
\usepackage{amsmath}
\usepackage{amsfonts}
\usepackage{amssymb} 
\usepackage{booktabs}
\usepackage{multirow}
\usepackage{subcaption}
\usepackage{algorithmic}
\usepackage{threeparttable}

\usepackage{comment}

\title{
SEED: Spectral Entropy-Guided Evaluation of Spatial-Temporal Dependencies for Multivariate Time Series Forecasting
}
\author{
    Feng Xiong\textsuperscript{\rm 1},  
    Zongxia Xie\textsuperscript{\rm 1 \footnote{Corresponding author}},  
    Yanru Sun\textsuperscript{\rm 1},
    Haoyu Wang\textsuperscript{\rm 2},  
    Jianhong Lin\textsuperscript{\rm 1}
}
\affiliations{
    \textsuperscript{\rm 1}Tianjin University\\
    \textsuperscript{\rm 2}Fudan University\\
    teddybear@tju.edu.cn, caddiexie@hotmail.com, yanrusun@tju.edu.cn, wanghy24@m.fudan.edu.cn, ljh123@tju.edu.cn
 
}

\usepackage{bibentry}

\begin{document}

\maketitle

\begin{abstract}
Effective multivariate time series forecasting often benefits from accurately modeling complex inter-variable dependencies. However, existing attention- or graph-based methods face three key issues: (a) strong temporal self-dependencies are often disrupted by irrelevant variables; (b) softmax normalization ignores and reverses negative correlations; (c) variables struggle to perceive their temporal positions.
To address these, we propose \textbf{SEED}, a Spectral Entropy-guided Evaluation framework for spatial-temporal Dependency modeling. SEED introduces a Dependency Evaluator, a key innovation that leverages spectral entropy to dynamically provide a preliminary evaluation of the spatial and temporal dependencies of each variable, enabling the model to adaptively balance Channel Independence (CI) and Channel Dependence (CD) strategies. To account for temporal regularities originating from the influence of other variables rather than intrinsic dynamics, we propose Spectral Entropy-based Fuser to further refine the evaluated dependency weights, effectively separating this part. Moreover, to preserve negative correlations, we introduce a Signed Graph Constructor that enables signed edge weights, overcoming the limitations of softmax. Finally, to help variables perceive their temporal positions and thereby construct more comprehensive spatial features, we introduce the Context Spatial Extractor, which leverages local contextual windows to extract spatial features.
Extensive experiments on 12 real-world datasets from various application domains demonstrate that SEED achieves state-of-the-art performance, validating its effectiveness and generality.
\end{abstract}

\begin{links}
    \link{Code}{https://github.com/saber1360/SEED}
\end{links}

\section{Introduction}

\begin{figure}[t!]
\centering
\includegraphics[width=1\columnwidth]{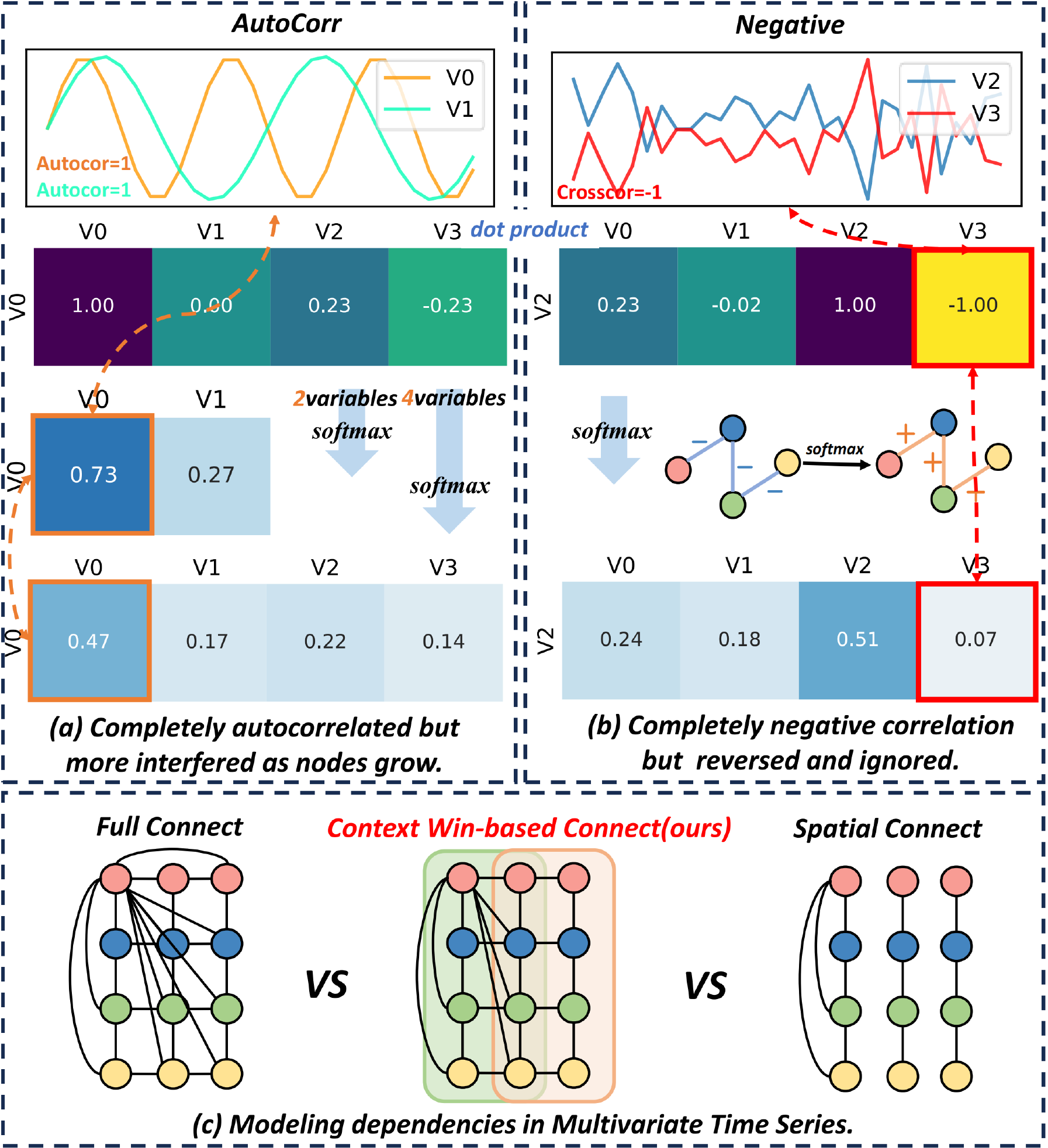}
\caption{
Challenges of temporal-spatial modeling. (a)show as the number of variables increases, they will be more influenced by other variables. (b) shows negative correlation in the time series, but is reversed by softmax function. (c) shows the method of modeling multivariate relationships. We employed local window modeling to enhance awareness of the temporal positions of spatial features.
}
\label{fig:figure1}
\end{figure}

Multivariate Time Series Forecasting (MTSF) is a fundamental yet highly challenging task, with broad applications in diverse domains such as financial market analysis~\cite{baffour2019hybrid}, traffic flow prediction~\cite{khan2023short}, energy consumption forecasting~\cite{chengqing2023multi}, and others~\cite{sun2021solar, sun2022accurate, wang2023drift}.
Recent advances in MTSF have primarily focused on attention-based~\cite{itransformer, crossformer, informer, sun2025learning} and graph-based~\cite{msgnet, shang2024ada, wang2024interdependency, luo2025hipatch} frameworks to model spatial-temporal relationships. The former connects all node pairs based on the similarity of the features, while the latter restricts the connection to only consider neighboring nodes that are close in distance. 



However, as shown in Figure~\ref{fig:figure1}, they still face the following three major challenges. \textbf{Firstly}, strong auto-correlation but as the number of nodes increases, the self-influence of these variables is weakened while interference from other external variables is amplified. In many datasets, some variables exhibit regular patterns and are predominantly influenced by their own historical behaviors. Recent strategies, such as Channel Partiality (CP)~\cite{qiu2025comprehensive}, aim to mitigate such interference by limiting inter-variable connections and reducing noise from unrelated channels~\cite{duet, chen2024similarity}. Although this helps alleviate the problem to some extent, it does not fundamentally address the issue, as interference remains significant with the increase in node count. \textbf{Secondly}, strong negative correlations are often reversed and ignored. Existing methods typically apply softmax normalization to the relationship matrix, resulting in all graph or attention weights being non-negative. This causes negative correlations between variables to be reversed or overlooked. Moreover, from the perspective of graph learning, enforcing non-negativity via softmax acts as a smoothing operation that retains only low-frequency components among node pairs~\cite{bo2021beyond}. In contrast, high-frequency components are also critical for learning node representations in heterogeneous graphs. Given the heterogeneity of variables in multivariate time series, existing graph construction methods fail to capture the diverse relationships among semantically different variables by using only non-negative scalar edge representations. \textbf{Finally}, the spatial structure cannot be modeled independently of the temporal context. Traditional spatial modeling methods typically consider only the current multivariate relationships~\cite{xue2023card} or use fully connected connections~\cite{yi2023fouriergnn}, the former ignored the temporal context, while the latter confused the temporal relationships. This naturally violate the real-world unified spatio-temporal dependencies, where the model lacks awareness of the temporal position of spatial features and cannot model continuous spatial-temporal interactions effectively.

To address the above challenges, we propose \textbf{SEED}, a Spectral Entropy-
guided evaluation framework for spatial-temporal dependency modeling. Specifically, we decouple spatio-temporal dependencies into temporal and spatial components and introduce a Spectral Entropy to perform an initial estimation of their weights. Subsequently, we design a Spectral Entropy-based Fuser (SE-Fuser) to further fuse these two. Spectral entropy effectively characterizes the energy concentration of time series in the frequency domain, reflecting the complexity of the time series. Time series with lower spectral entropy have stronger regularity and periodicity, thus guiding each variable in balancing between channel-independent (CI) and channel-dependent (CD) strategies. For series with strong structures, more attention should be given to their own temporal dynamics, while more random series are more influenced by other variables and should focus on complex spatial dependencies.
In addition, we design two Signed Graph Constructor to capture negative correlations between variables, one based on \emph{tanh}, which provides a more dispersed weight distribution, and the other based on \emph{softmax}, which yields more concentrated weights. Furthermore, inspired by the theories of multi-graph and signed graph modeling, we adopt a multi-head mechanism to capture richer and more diverse inter-variable relationships. Finally, to unify spatial and temporal perspectives and enhance complementarity, we introduce Context Spatial Extractor to construct more comprehensive spatial features that are aware of their temporal positions.
Our main contributions are summarized as follows:
\begin{itemize}
    \item We propose a spatial-temporal feature fusion method that reasonably balances CI and CD while preserving internal structures of individual variables.
    \item We introduce \textbf{SEED}, a novel spatial-temporal model that accounts for each variable's specific attention to spatial-temporal features, and allows edges in graph to be represented by signed vectors to model complex and diverse inter-variable relationships.
    \item Extensive experiments demonstrate that SEED achieves state-of-the-art performance on multiple widely-used benchmark datasets.
\end{itemize}

\section{Related Work}
\subsection{Multivariate Time Series Forecasting Models}
Multivariate Time Series Forecasting (MTSF) is a key area in time series analysis~\cite{ma2024survey, qiu2024tfb, sun2025ppgf}. With the strong representation power of neural networks, deep learning has shown great potential in this field~\cite{wang2025chattime, sun2025hierarchical}. Existing MTSF models mainly focus on capturing inter-variable dependencies. CNN-based~\cite{zhou2025crosslinear, zhou2021informer, wu2021autoformer} and MLP-based~\cite{ekambaram2023tsmixer} methods model variable interactions adaptively, while Transformer variants employ attention for dynamic aggregation. GNN-based approaches~\cite{yi2023fouriergnn, zhao2023multiple, wang2024fully} (e.g., GCNs, GATs) construct graphs to capture topological dependencies. Yet, some studies find that models without explicit inter-variable modeling can still perform well, likely because certain variables follow simpler dynamics, where complex inter-variable modeling is unnecessary for good predictions. Conversely, complex variables benefit from richer relational modeling. These findings inspire us to integrate both modeling philosophies for balanced performance.

\subsection{Spatial-Temporal Dependencies Modeling}
In multivariate time series forecasting (MTSF), dependency modeling methods can be broadly categorized into three types: Channel-Independent (CI) strategies, Channel-Dependent (CD) strategies, and Channel Partiality (CP) strategies. CI strategies focus on temporal dependencies, typically resulting in simpler and more robust models~\cite{sun2025learning, liu2025learning}. However, they often perform poorly when dealing with complex dynamics or high-dimensional data. CD strategies introduce explicit modeling of inter-variable dependencies, emphasizing spatial relationships~\cite{wu2025catch}. While this can provide useful information gain, it often increases model complexity and may introduce redundant information in tasks where self-dependence dominates. Recently, research has shifted toward Channel Partiality, which attempts to selectively model variable dependencies either statically or dynamically. However, such methods heavily rely on feature-level similarity for clustering and often lack consideration of the inherent periodic patterns within the time series itself. To address this, we propose to leverage spectral entropy of the series as a dynamic criterion to balance the CI and CD strategies.


\begin{figure*}[t!]
\centering
\includegraphics[width=0.98\textwidth]{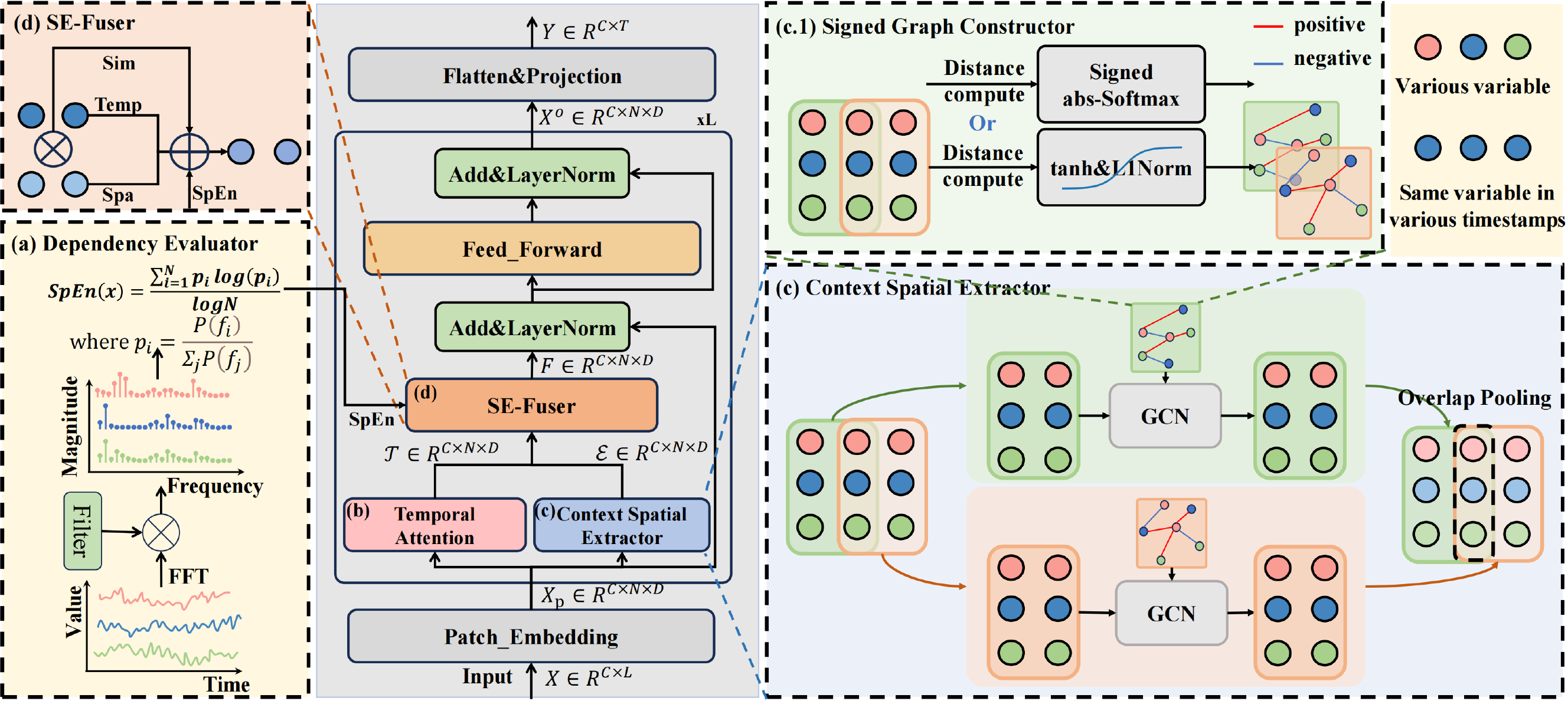} 
\caption{Overview of SEED, which comprises the following key modules:  (a) \textbf{Dependency Evaluator} calculates the spectral entropy for evaluation of Spatial-Temporal Dependencies after filtering in the frequency domain; (b) \textbf{Temporal Attention} focuses on utilize its own information to capture own internal patterns; (c) \textbf{Context Spatial Extractor} extracts the more complete spatial feature from the surrounding spatial-temporal context; (c.1) \textbf{Signed Graph Constructor} allows the edge weights between nodes to be negative weights; (d) \textbf{SE-Fuser} dynamically fuses the variable's temporal feature with spatial feature.
}
\label{fig:overview}
\end{figure*}

\section{Preliminary}
\subsection{Problem Formulation}

In the context of multivariate time series forecasting, let $\mathbf{X=\{x_1,x_2,...,x_C\}}\in\mathbb{R}^{C\times L}$ be the input time series, where $C$ denotes the number of channels and $L$ denotes the length of the lookback window. $\mathbf{x_i}\in\mathbb{R}^{L}$ represents one of the channels. The objective is to evaluate the future values $\hat{X} \in \mathbb{R}^{C \times T}$, where $T$ denotes the length of the forecasting window. The forecast values are given by $\mathbf{\hat{Y}=\{\hat{y}_{1},\hat{y}_{2},...,\hat{y}_{C}\}}\in\mathbb{R}^{C\times T}$, where $T$ denotes the forecasting horizon.

\section{Method}
\subsection{Overall Architecture}
The comprehensive architecture of SEED, as illustrated in Figure \ref{fig:overview}, comprised of five key components: 
\textbf{(a) Dependency Evaluator Module}(DpdEva) measures the relatively complex and random parts of variables by the spectral entropy after filtering; 
\textbf{(b) Temporal Attention Module}(TAttn) focuses on capturing the relational patterns within the variable itself; 
\textbf{(c) Context Spatial Extractor Module}(CSE) extracts the more complete spatial feature from the surrounding spatial-temporal context; 
\textbf{(c.1) Signed Graph Constructor Module}(SGC) constructs a non-symmetric graph with signed vector-valued edges, using a multi-head mechanism and adopting a relaxed learnable distance metric.
\textbf{(d) Spectral Entropy-based Fuser Module}(SE-Fuser) considers the extent of the involvement of spatial feature in the the inherent regular patterns exhibited by the variables and fuses the temporal feature with spatial feature.

We first evaluate the inherent regular patterns of each variable in the input \( X \in \mathbb{R}^{C \times L} \) by using spectral entropy computed after filtering.
\begin{equation}
\text{SpEn} = \text{DpdEva}(X).
\end{equation}
The resulting $\text{SpEn}\in\mathbb{R}^{C}$ represents the spectral entropy, which is used to measure the inherent regular patterns of each variable.

We then divide the input multivariate time series \( X \) into non-overlapping patches $\mathbf{X}_p^{'}\in\mathbb{R}^{C\times N\times P}$, where $P$ denotes length of each patch and $N=\lceil\frac{L}{P}\rceil$ denotes the number of patches. Then each patch is mapped to an embedded patch token $\mathbf{X}_p\in\mathbb{R}^{C\times N\times D}$.
\begin{align}
    \mathbf{X}_p^{'}=\text{Patching}(X), \mathbf{X}_p=\text{Embedding}(\mathbf{X}_p^{'})+\text{PE},
\end{align}
where $\text{PE}$ represents the positional encoding, which is added to the embedded patches to retain the directional information and the relative positions among patches.


The patch token $\mathbf{X}_p$ is then concurrently fed into two distinct pathways:
\begin{align}
\mathcal{T} = \text{TAttn}(\mathcal{X}^{(l-1)}), \mathcal{E} = \text{CSE}(\mathcal{X}^{(l-1)}),
\end{align}
where $\text{TAttn}$(·) and $\text{CSE}$(·) are the Temporal Attention module and Context Spatial Extractor module, respectively. The output $ \mathcal{T}\in\mathbb{R}^{C\times N\times D} $ denotes temporal feature and $ \mathcal{E}\in\mathbb{R}^{C\times N\times D} $ denotes the spatial feature.

After that, the resulting $\mathcal{T}$ and $\mathcal{E}$ are fed into the SE-Fuser module and fused based on the $\text{SpEn}$.

\begin{equation}
\mathcal{F} = \text{SE-Fuser}(\mathcal{T}, \mathcal{E}, \text{SpEn}).
\end{equation}

Subsequently, the output is passed through a feed-forward MLP block and skip connection:
\begin{align}
\mathcal{H}^{(l)} &= \text{LayerNorm}(\mathcal{X}^{(l-1)} + \mathcal{F}), \\
\mathcal{X}^{(l)} &= \text{LayerNorm}(\mathcal{H}^{(l)} + \text{Linear}(\mathcal{H}^{(l)})).
\end{align}

Finally, we flatten the patch-level representations of each variable and project them to the output dimension:
\begin{equation}
\hat{Y} = \text{Projection}(\text{FlattenHead}(\mathcal{X}^{(o)})),
\end{equation}
where \( \hat{Y} \in \mathbb{R}^{C \times T} \) denotes the forecasted series, and \( T \) is the forecasting horizon.

\subsection{Dependency Evaluator Module}

The Dependency Evaluator Module aims to quantify how much of each variable's temporal dynamics can be confidently predicted using only its own historical information. To this end, we employ \textbf{spectral entropy} as a frequency-domain measure of structural complexity. Detailed spectral entropy analysis is provided in Appendix. However, given the sensitivity of spectral entropy to noise, we incorporate frequency-domain filtering to eliminate noise disturbances and improve robustness.





Given an input time series \( x_c \in \mathbb{R}^{L} \) for variable \( c \), we first apply the Fourier Transform and a Plain Shaping Filter~\cite{yi2024filternet} to obtain a denoised frequency spectrum:
\begin{eqnarray}\label{eq_arbfilter}
    \mathcal{Z}_c=\mathcal{F}(\mathbf{x_c}),  \mathcal{S}_c = \mathcal{Z}_c \odot_{L} \mathcal{H}_{\phi},
\end{eqnarray}
where \( \mathcal{F} \) denotes the Fourier Transform, \( \odot_{L} \) is the element-wise multiplication along the length dimension \( L \), \( \mathcal{H}_{\phi} \in \mathbb{C}^{1 \times L} \) is the universal plain shaping filter, and \( \mathcal{S}_c \in \mathbb{C}^{1 \times L} \) is the filtered spectrum.

We then compute the Power Spectral Density (PSD) of variable \( c \) as:
\begin{equation}
P_c(f_i) = |\mathcal{S}_c(f_i)|^2, \quad i = 1, 2, \dots, L,
\end{equation}
where $P_c(f_i)$ is the power at frequency bin $f_i$ for variable $c$.

Finally, we calculate the spectral entropy of \( x_c \) as:
\begin{equation}
\text{SpEn}(x_c) = \frac{- \sum_{i=1}^{L} p_{c,i} \log(p_{c,i}) }{\log(L)},
\end{equation}
where \( p_{c,i} = \frac{P_c(f_i)}{\sum_{j=1}^{L} P_c(f_j)} \) is the normalized power at frequency bin \( f_i \), forming a valid probability distribution over frequencies.


The resulting spectral entropy \( \text{SpEn}(x_c) \in [0, 1] \) quantifies the level of uncertainty or disorder in the frequency domain. A lower SpEn value indicates that the spectral energy is more concentrated, implying stronger temporal regularity and higher self-prediction ability. In such cases, the variable is considered to be more self-reliable, and thus self-modeling is more trustworthy. Conversely, a higher SpEn value suggests that the energy is more dispersed across frequencies, reflecting greater complexity and stochasticity, which requires the model to rely more on external contextual information gain for accurate forecasting. 

We apply this computation to all variables \( c = 1, \dots, C \), resulting in a spectral entropy vector \( \text{SpEn} \in \mathbb{R}^C \).

\subsection{Temporal Attention Module}

The Temporal Attention Module is designed to model intravariable dependencies across time, capturing how each variable evolves based on its own history. Similar to PatchTST~\cite{patchtst}, we apply temporal multi-head attention to each variable independently.

Given the embedded patch tokens \( \mathbf{X}_p \in \mathbb{R}^{C \times N \times D} \), we compute temporal attention for each variable \( c \) as:
\begin{align}
Q_c, K_c, V_c &= \text{Linear}(\mathbf{X}_{p,c}), \\
\text{Attn}_c &= \text{Softmax}\left(\frac{Q_c K_c^\top}{\sqrt{d_k}}\right) V_c, \\
\mathcal{T}_c &= \text{Concat}(\text{head}_1, \dots, \text{head}_h) W^O,
\end{align}
where \( d_k \) is the dimension of each head, \( h \) is the number of attention heads, and \( W^O \) is a learned projection matrix. The output \( \mathcal{T} \in \mathbb{R}^{C \times N \times D} \) preserves temporally structured representations for each variable.

\subsection{Context Spatial Extractor Module}

To capture the complex and dynamic inter-variable dependencies, we design an Context Spatial Extractor module, which constructs local spatial-temporal interaction graphs from embedded patches and models the signed relationships across variables.

Given the embedded feature series $X_p \in \mathbb{R}^{C \times N \times D}$, where $C$ is the number of variables, $N$ is the number of patches per variable, and $D$ is the feature dimension of each patch, we first partition the temporal dimension $N$ into overlapping local windows of size 2 with stride 1. This results in a local patch tensor $X_{win} \in \mathbb{R}^{C \times (N{-}1) \times 2 \times D}$. Such a formulation allows the spatial interactions among variables to be aware of their temporal positions, achieving unified spatial-temporal perception.

\subsubsection{ Signed Graph Constructor Module}

For $k$-th local window $X_{win}^k \in \mathbb{R}^{C \times 2 \times D}$, we flatten the first two dimensions to obtain a set of spatial-temporal patches with $n=C\times 2$ patches containing the enhanced local information $\mathbf{X}_{local}^k\in\mathbb{R}^{n\times D}$. 

To model more diverse relationships among variables, we, inspired by multi-graph theory, signed graphs, attention mechanisms, divided the features into multiple heads, allowing the edges in the inter-variable relationship graph to be represented as a set of signed vectors.Specifically, we calculate the projection distances in a multi-head mechanism, allowing for modeling more complex relationships. We first divide $\mathbf{X}_{local}^k$ into different heads $\mathbf{X}_h^k\in\mathbb{R}^{H\times n\times \lfloor\frac{D}{H}\rfloor}$ and then calculate distances $\text{Dist}(\cdot)$.
Here, $H$ is the number of heads and $A^T$ is the transpose.
\begin{equation}
    \text{Dist}(\mathbf{X}_h^k)=(\mathbf{X}_h^k)^T\mathbf{Q}\mathbf{X}_h^k\in\mathbb{R}^{H\times n\times n},
\end{equation}
where $ \mathbf{Q} \in \mathbb{R}^{n \times n}$ denote a learnable matrix. 


Next, based on above distances, we construct the signed graph. Considering that Softmax tends to highlight prominent associations, whereas methods for constructing inter-variable relationships based on CNN/MLP often result in more uniform relationships, we designed two distinct signed graph Constructor methods: one based on Softmax and another based on Tanh to explore these differences.

Specifically, we construct a signed interaction graph using two alternative strategies:

Let $s_{ij}$ denotes the edge connecting variable i and variable j in distance $ \text{Dist}(\mathbf{X}_h^k) $.
\begin{itemize}
    \item \textbf{softmax-Based Signed Graph:} We then decouple the sign and magnitude: the magnitude $|s_{ij}|$ is used to compute weights via softmax normalization, while the sign is preserved and multiplied with the attention, leading to the signed attention matrix:
    \begin{equation}
        \alpha_{ij} = \text{sign}(s_{ij}) \cdot \frac{\exp(|s_{ij}|)}{\sum_{j'=1}^{n} \exp(|s_{ij'}|)} .
    \end{equation}
   
    \item \textbf{tanh-Based Signed Graph:} Alternatively, we compute cosine similarity followed by a $\tanh$ transformation to retain bounded signed values. The resulting matrix is normalized using the L1 norm:
    \begin{equation}
        \alpha_{ij} = \frac{\tanh(s_{ij})}{\sum_j |\tanh(s_{ij})|} .
    \end{equation}
\end{itemize}
The absolute value of the above matrix is finally used as the criterion for construction of $k$-nearest neighbors (KNN), resulting in a set of graphs $\mathbf{G} \in \mathbb{R}^{ H \times n \times n} $.

\subsubsection{ GCN and Overlap Pooling }
For the features of the $k$-th patch, we consider windows $\mathbf{X}_{\text{win}}^{k-1}$ and $\mathbf{X}_{\text{win}}^{k}$.
\begin{align}
E_\text{k-1} = \text{GCN}(\mathbf{X}_{\text{win}}^{k-1},\mathbf{G}^{k-1}), 
E_\text{k} = \text{GCN}(\mathbf{X}_{\text{win}}^{k},\mathbf{G}^k),
\end{align}
where $\mathbf{E}_\text{k} \in \mathbb{R}^{ C \times 2 \times D}$
These two are then aggregated (e.g., via mean or max pooling):
\begin{equation}
\mathcal{E}_k = \text{OverlapPooling}(E_\text{k-1}, E_\text{k}),
\end{equation}
where $\text{OverlapPooling} $ denotes pooling the overlapping parts from the preceding and succeeding windows, and $ \mathcal{E}_k \in \mathbb{R}^{ C \times 1 \times D} $ denotes the features of the $k$-th patch.
We apply this computation to all patch \( k = 1, \dots, N \), resulting in a vector \( \mathcal{E} \in \mathbb{R}^{ C \times N \times D} \).

\subsection{Spectral Entropy-based Fuser  Module}

The Spectral Entropy-based Fuser (SE-Fuser) integrates the temporal features and spatial features based on the confidence signal from the DpdEva module. The fusion is guided by both patch-wise feature similarity and spectral entropy.

For each variable \( c \), we compute a fusion coefficient \( \alpha_c \in [0, 1] \) based on its spectral entropy:
\begin{equation}
\alpha_c = 1-\text{SpEn}(x_c),
\end{equation}
where \( \text{SpEn}(x_c) \) denotes spectral entropy of variable c.

Let \( \text{Sim}_{c,n} \) denote the similarity between temporal features \( \mathcal{T}_{c,n} \) and spatial features \( \mathcal{E}_{c,n} \), where c denote variable c and n denote n-th patch. The fusion weight becomes:
\begin{equation}
w_{c,n} = \alpha_c \cdot (1-\text{Sim}_{c,n}).
\end{equation}
When $\text{Sim}_{c,n}$ is low, it suggests that the regularity exhibited by the variable has a weaker association with the spatial features, in which case we can appropriately strengthen the utilization of temporal features. Conversely, when $\text{Sim}_{c,n}$ is high, it indicates that the variable's regularity relies more heavily on the spatial features, and we should reduce the reliance on temporal features.

Then the final fused representation is computed as a weighted combination:
\begin{equation}
\mathcal{F}_{c,n} = w_{c,n} \cdot \mathcal{T}_{c,n} + (1-w_{c,n}) \cdot \mathcal{E}_{c,n}.
\end{equation}
This allows the model to rely more on intrinsic temporal dynamics when the variable exhibits strong internal regularities, and to shift toward more complex spatial modeling when the variable exhibits higher complexity or uncertainty.

\subsection{Loss Function}

The loss function of SEED consists of two components.
For the predictors, the Mean Squared Error (MSE) loss is used to measure the variance between predicted values and ground truth.
\begin{equation}
    \boldsymbol{\mathcal{L}}_\text{pred}=\frac{1}{T}\sum_{i=0}^{T}\|\mathbf{y_{:,i}-\hat{y}_{:,i}}\|_{2}^{2}.
\end{equation}

Moreover, we introduce an Loss based on spectral entropy $\boldsymbol{\mathcal{L}}_\text{SpEn}$ to Guide the more reasonable distribution of weights.
\begin{equation}
    \boldsymbol{\mathcal{L}}_\text{SpEn}=\frac{1}{C}\sum_{c=1}^{C}\|\mathbf{\text{SpEn}(y_{c,:})-\text{SpEn}(\hat{y}_{c,:})}\|_{2}^{2}.
\end{equation}

Therefore, the final loss function is defined as:
\begin{equation}
 \boldsymbol{\mathcal{L}}=\boldsymbol{\mathcal{L}}_\text{pred}+
 \lambda \boldsymbol{\mathcal{L}}_\text{SpEn},
\end{equation}
where $\lambda$  are the scaling factors.

\begin{table*}[!htb]
\centering
\scalebox{0.9}{
\begin{tabular}{c|cc|cc|cc|cc|cc|cc}
\toprule
\multicolumn{1}{c}{\multirow{2}{*}{\scalebox{1.1}{Models}}}
& \multicolumn{2}{c}{SEED}
& \multicolumn{2}{c}{TQNet}
& \multicolumn{2}{c}{DUET}
& \multicolumn{2}{c}{iTransformer}
& \multicolumn{2}{c}{MSGNet}
& \multicolumn{2}{c}{SOFTS} \\

\multicolumn{1}{c}{}
& \multicolumn{2}{c}{\scalebox{0.8}{(\textbf{Ours})}}
& \multicolumn{2}{c}{\scalebox{0.8}{CD ~\citeyearpar{tqnet}}}
& \multicolumn{2}{c}{\scalebox{0.8}{CP ~\citeyearpar{duet}}}
& \multicolumn{2}{c}{\scalebox{0.8}{CD ~\citeyearpar{itransformer}}}
& \multicolumn{2}{c}{\scalebox{0.8}{CD ~\citeyearpar{msgnet}}}
& \multicolumn{2}{c}{\scalebox{0.8}{CD~\citeyearpar{softs}}} \\

\cmidrule(lr){2-3} \cmidrule(lr){4-5} \cmidrule(lr){6-7} \cmidrule(lr){8-9} \cmidrule(lr){10-11} \cmidrule(lr){12-13}

\multicolumn{1}{c}{Metric}
& \scalebox{0.85}{MSE} & \scalebox{0.85}{MAE}
& \scalebox{0.85}{MSE} & \scalebox{0.85}{MAE}
& \scalebox{0.85}{MSE} & \scalebox{0.85}{MAE}
& \scalebox{0.85}{MSE} & \scalebox{0.85}{MAE}
& \scalebox{0.85}{MSE} & \scalebox{0.85}{MAE}
& \scalebox{0.85}{MSE} & \scalebox{0.85}{MAE} \\

\toprule

ETTm1 & \textbf{0.369} & \textbf{0.391} & \underline{0.377} & \underline{0.393} & 0.390 & \underline{0.393} & 0.407 & 0.410 & 0.398 & 0.411 & 0.393 & 0.403 \\

\midrule

ETTm2 & \textbf{0.273} & \textbf{0.321} & \underline{0.277} & \underline{0.323} & 0.280 & 0.324 & 0.288 & 0.332 & 0.288 & 0.330 & 0.287 & 0.330 \\

\midrule

ETTh1 & \textbf{0.418} & \textbf{0.427} & \underline{0.441} & \underline{0.434} & 0.443 & 0.436 & 0.454 & 0.447 & 0.452 & 0.452 & 0.449 & 0.442 \\

\midrule

ETTh2 & \textbf{0.365} & \textbf{0.396} & 0.378 & 0.402 & \underline{0.372} & \underline{0.397} & 0.383 & 0.407 & 0.396 & 0.417 & 0.385 & 0.408 \\

\midrule

Weather & \textbf{0.239} & \underline{0.270} & \underline{0.242} & \textbf{0.269} & 0.251 & 0.273 & 0.258 & 0.279 & 0.249 & 0.278 & 0.255 & 0.278 \\

\midrule

ECL & \textbf{0.156} & \textbf{0.252} & \underline{0.164} & 0.259 & 0.172 & \underline{0.258} & 0.178 & 0.270 & 0.194 & 0.300 & 0.174 & 0.264 \\

\midrule

Traffic & \textbf{0.404} & \textbf{0.265} & 0.445 & 0.276 & 0.451 & 0.269 & 0.428 & 0.282 & 0.641 & 0.370 & \underline{0.409} & \underline{0.267} \\

\midrule

Solar & \underline{0.218} & \underline{0.252} & \textbf{0.198} & 0.256 & 0.237 & \textbf{0.233} & 0.233 & 0.262 & 0.262 & 0.288 & 0.229 & 0.256 \\

\midrule

{1st Cnt} & \textbf{7} & \textbf{6} & 1 & 1 & 0 & 1 & 0 & 0 & 0 & 0 & 0 & 0 \\

\bottomrule
\end{tabular}
}
\caption{Results of the multivariate long-term time series forecasting task, evaluated using MSE and MAE (lower is better). The input series length $L$ is set to $96$ for all baselines. The best results are highlighted in \textbf{bold}, while the second-best results are \underline{underlined}. See Appendix for full results.}
\label{tab::long_result_cut}
\end{table*}

\begin{table*}[!htb]
\centering
\scalebox{0.9}{
\begin{tabular}{c|cc|cc|cc|cc|cc|cc}
\toprule
\multicolumn{1}{c}{\multirow{2}{*}{\scalebox{1.1}{Models}}}
& \multicolumn{2}{c}{SEED}
& \multicolumn{2}{c}{iTransformer}
& \multicolumn{2}{c}{Leddam}
& \multicolumn{2}{c}{SOFTS}
& \multicolumn{2}{c}{PatchTST}
& \multicolumn{2}{c}{Crossformer} \\
\multicolumn{1}{c}{}
& \multicolumn{2}{c}{\scalebox{0.8}{(\textbf{Ours})}}
& \multicolumn{2}{c}{\scalebox{0.8}{CD ~\citeyearpar{itransformer}}}
& \multicolumn{2}{c}{\scalebox{0.8}{CD ~\citeyearpar{leddam}}}
& \multicolumn{2}{c}{\scalebox{0.8}{CD ~\citeyearpar{softs}}}
& \multicolumn{2}{c}{\scalebox{0.8}{CI ~\citeyearpar{patchtst}}}
& \multicolumn{2}{c}{\scalebox{0.8}{CD~\citeyearpar{crossformer}}} \\

\cmidrule(lr){2-3} \cmidrule(lr){4-5} \cmidrule(lr){6-7} \cmidrule(lr){8-9} \cmidrule(lr){10-11} \cmidrule(lr){12-13}

\multicolumn{1}{c}{Metric}
& \scalebox{0.85}{MSE} & \scalebox{0.85}{MAE}
& \scalebox{0.85}{MSE} & \scalebox{0.85}{MAE}
& \scalebox{0.85}{MSE} & \scalebox{0.85}{MAE}
& \scalebox{0.85}{MSE} & \scalebox{0.85}{MAE}
& \scalebox{0.85}{MSE} & \scalebox{0.85}{MAE}
& \scalebox{0.85}{MSE} & \scalebox{0.85}{MAE} \\

\toprule








PEMS03 & \textbf{{0.084}} & \textbf{{0.188}} & 0.096 & 0.204 & 0.101 & 0.210 & \underline{{0.087}} & \underline{{0.192}} & 0.151 & 0.265 & 0.138 & 0.253 \\

\midrule
PEMS04 & \textbf{{0.080}} & \textbf{{0.182}} & 0.098 & 0.207 & 0.102 & 0.213 & \underline{{0.091}} & \underline{{0.196}} & 0.162 & 0.273 & 0.145 & 0.267 \\

\midrule

PEMS07 & \textbf{{0.068}} & \textbf{{0.165}} & 0.088 & 0.190 & 0.087 & 0.192 & \underline{{0.075}} & \underline{{0.173}} & 0.166 & 0.270 & 0.181 & 0.272 \\

\midrule

PEMS08 & \textbf{{0.080}} & \textbf{{0.182}} & 0.127 & 0.212 & \underline{{0.102}} & 0.211 & 0.114 & \underline{{0.208}} & 0.238 & 0.289 & 0.232 & 0.270 \\

\midrule

{1st Cnt} & \textbf{4} & \textbf{4} & 0 & 0 & 0 & 0 & 0 & 0 & 0 & 0 & 0 & 0 \\

\bottomrule

\end{tabular}}

\caption{Short-term forecasting results. The input length $L$ is $96$. All results are averaged across three different forecasting horizons: $T \in \{12, 24, 48\}$. The best results are highlighted in \textbf{bold}, while the second-best results are \underline{underlined}. See Appendix for full results.}
\label{tab::short_result_cut}
\end{table*}

\section{Experiments}
\subsection{Experimental Details}
\subsubsection{Datasets.}

For long-term forecasting, we conduct extensive experiments on $8$ widely-recognized multivariate time series forecasting datasets, including ETT (ETTh1, ETTh2, ETTm1, ETTm2), Traffic, Electricity, Weather, Solar-Energy~\cite{miao2024less,miao2024unified} datasets. For short-term forecasting, we  selected $4$ benchmarks from PEMS (PEMS03, PEMS04, PEMS07, PEMS08)~\cite{liu2025timecma,liu2025timekd}. The statistics of the dataset are shown in Appendix. 

\subsubsection{Baselines.}

We compare our method with SOTA representative methods, including 
GNN-based methods: MSGNet~\cite{msgnet}; 
Transformer-based methods: iTransformer~\cite{itransformer}, TQNet~\cite{tqnet};
, Crossformer~\cite{crossformer}
Linear-based methods: Leddam~\cite{leddam}, SOFTS~\cite{softs}, DUET~\cite{duet};

\subsubsection{Setup.}
All experiments are implemented in PyTorch and conducted on workstations equipped with 8 NVIDIA GeForce RTX 3090 24GB GPUs. We adopted a consistent experimental setup identical to that of iTransformer ~\cite{itransformer} to ensure a fair comparison.  Specifically, the lookback length for all models was fixed at 96, and the SGC Module module employs a tanh-Based Signed Graph. Mean Absolute Error (MAE) and Mean Squared Error (MSE) was used as metrics.

\subsection{Main Results}

\subsubsection{Long-term Forecasting.}
Table \ref{tab::long_result_cut} presents the predictive performance of SEED in multivariate long-term time series forecasting tasks in eight datasets. he input length $L$ is $96$ for our method and all baselines. The forecasting horizon $T$ is $\{96,192,336,720\}$. From the table, SEED demonstrates superior accuracy over current state-of-the-art models in most cases.  Specifically, averaging MSE across all prediction lengths, SEED achieved the best performance on 6 out of the 8 datasets and secured the second-best performance on the remaining 2 datasets. 

\subsubsection{Short-term Forecasting.}
Table \ref{tab::short_result_cut} presents the predictive performance of SEED in multivariate short-term time series forecasting tasks in eight datasets, The input length $L$ is $96$ for our method and all baselines. The forecasting horizon $T$ is $\{12,24,48\}$. From the table, SEED consistently outperforms other methods across all 4 PEMS datasets. 
The insignificant improvement in SEED performance in PEMS03 and PEMS07 might be attributed to the high number of variables in these datasets, resulting in the temporal-dependent part having a relatively smaller effect.

\begin{table*}[t!]
\centering
\scalebox{0.9}{
\begin{tabular}{c|cc|cc|cc|cc|cc}
\toprule
\multirow{2}{*}{\centering\scalebox{1.2}{\textbf{Variants}}} & \multicolumn{2}{c|}{\textbf{ETT (avg)}} & \multicolumn{2}{c|}{\textbf{Weather}} & \multicolumn{2}{c|}{\textbf{ECL}} & \multicolumn{2}{c|}{\textbf{Traffic}} & \multicolumn{2}{c}{\textbf{Solar-Energy}} \\
\cmidrule(lr){2-3} \cmidrule(lr){4-5} \cmidrule(lr){6-7} \cmidrule(lr){8-9} \cmidrule(lr){10-11}
& MSE & MAE & MSE & MAE & MSE & MAE & MSE & MAE & MSE & MAE \\
\toprule

\textbf{w/o TAttn} 
& 0.364 & 0.387 & 0.244 & 0.272 & 0.159 & 0.254 & 0.407 & 0.268 & 0.230 & 0.257 \\
\textbf{w/o CSE} 
& 0.363 & 0.388 & 0.245 & 0.274 & 0.178 & 0.268 & 0.437 & 0.274 & 0.240 & 0.269 \\
\midrule

\textbf{re-S1} 
& 0.360 & 0.386 & 0.243 & 0.272 & 0.177 & 0.272 & 0.411 & 0.257 & 0.226 & 0.253 \\
\textbf{re-S2}
& 0.365 & 0.389 & 0.242 & 0.271 & 0.162 & 0.257 & 0.411 & 0.270 & 0.223 & 0.252 \\
\midrule
\textbf{re-F1} 
& 0.362 & 0.387 & 0.242 & 0.272 & 0.161 & 0.257 & 0.407 & 0.269 & 0.231 & 0.260 \\
\textbf{re-F2} 
& 0.363 & 0.388 & 0.245 & 0.273 & 0.160 & 0.255 & 0.408 & 0.269 & 0.227 & 0.256 \\
\textbf{re-F3}
& 0.374 & 0.392 & 0.374 & 0.392 & 0.170 & 0.256 & 0.235 & 0.300 & 0.297 & 0.340 \\
\midrule
\textbf{re-C1}
& 0.366 & 0.390 & 0.242 & 0.273 & 0.167 & 0.262 & 0.413 & 0.270 & 0.229 & 0.258 \\
\textbf{re-C2}
& 0.363 & 0.388 & 0.240 & 0.270 & 0.164 & 0.260 & 0.413 & 0.269 & 0.231 & 0.255 \\

\midrule
\textbf{SEED} 
& \textbf{0.356} & \textbf{0.384} & \textbf{0.239} & \textbf{0.270} & \textbf{0.156} & \textbf{0.252} & \textbf{0.404} & \textbf{0.265} & \textbf{0.218} & \textbf{0.252} \\
\bottomrule
\end{tabular}}
\caption{Ablation analysis showing averaged ETT results and other datasets. results. All results are averaged across three different forecasting horizons: $T \in \{96, 192, 336, 720\}$. The best results are highlighted in \textbf{bold}. See Appendix for full results.
}
\label{tab:ablation_avg}
\end{table*}

\subsection{Ablation Study}
To validate the effectiveness of SEED, we conduct a  comprehensive ablation study on its architectural design. 

\subsubsection{Spatial-Temporal Module Ablation.}
\textbf{w/o-TAttn}: The Temporal Attention Module was removed.
\textbf{w/o-CSE}: The Context Spatial Extractor Module was removed.

\subsubsection{Replace of SGC Module.}
The Signed Graph Constructor Module in the aforementioned experiment uses the tanh-based signed graph. Its construction method ensures that the absolute values of the weights in the graph are relatively smooth. Therefore, we further explored the differences between this construction method, which highlights key points using softmax, and the previous one.
\textbf{re-S1}: The Signed Graph Constructor Module has been replaced by original softmax. That is, to ignore the negative correlation.
\textbf{re-S2}: The Signed Graph Constructor Module use softmax-Based Signed Graph.

\subsubsection{Replace of SE-Fuser Module.}
In order to verify the effectiveness of our spatiotemporal feature fusion method, we designed the following three variants:
\textbf{re-F1}: Replace the spectral entropy with learnable weights as the weights.
\textbf{re-F2}: The cross-use of the TAttn and CSE  Module.
\textbf{re-F3}: Concatenate the two features and pass them through a linear layer as the fusion weights. 

\subsubsection{Replace of CSE Module.}
We designed the following two spatial modeling methods to replace our CSE module.
\textbf{re-C1}: Only consider the relationships between different variables at the same moment.
\textbf{re-C2}: Use the full connection structure with all patches.

\subsection{Model Analysis}
\subsubsection{Case study on Signed Graph modeling.}
\begin{figure}
    \centering
    \includegraphics[width=1.0\linewidth]{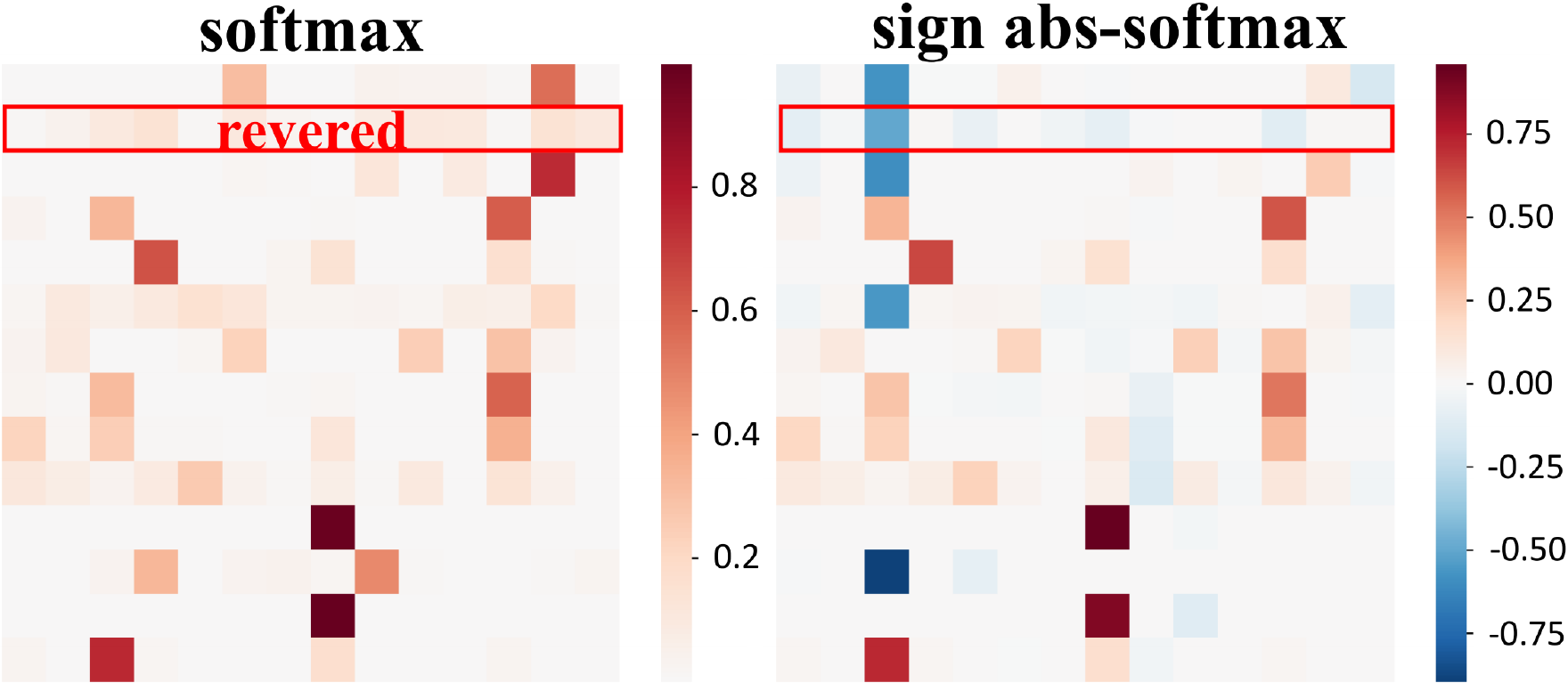}
    \caption{Above are the dependencies learned by SEED from the ETTm1. left: softmax function; right: our softmax-based signed graph.}
    \label{fig:signed_graph}
\end{figure}

As shown in  Figure \ref{fig:signed_graph}, we visualize the dependencies of selected filters of one batch in the ETTm1. In fact, it shows that constructing the graph using softmax may lead to a group of negatively correlated connections being reversed into positive ones.


\subsubsection{Model Efficiency}
\begin{figure}
    \centering
    \includegraphics[width=1.0\linewidth]{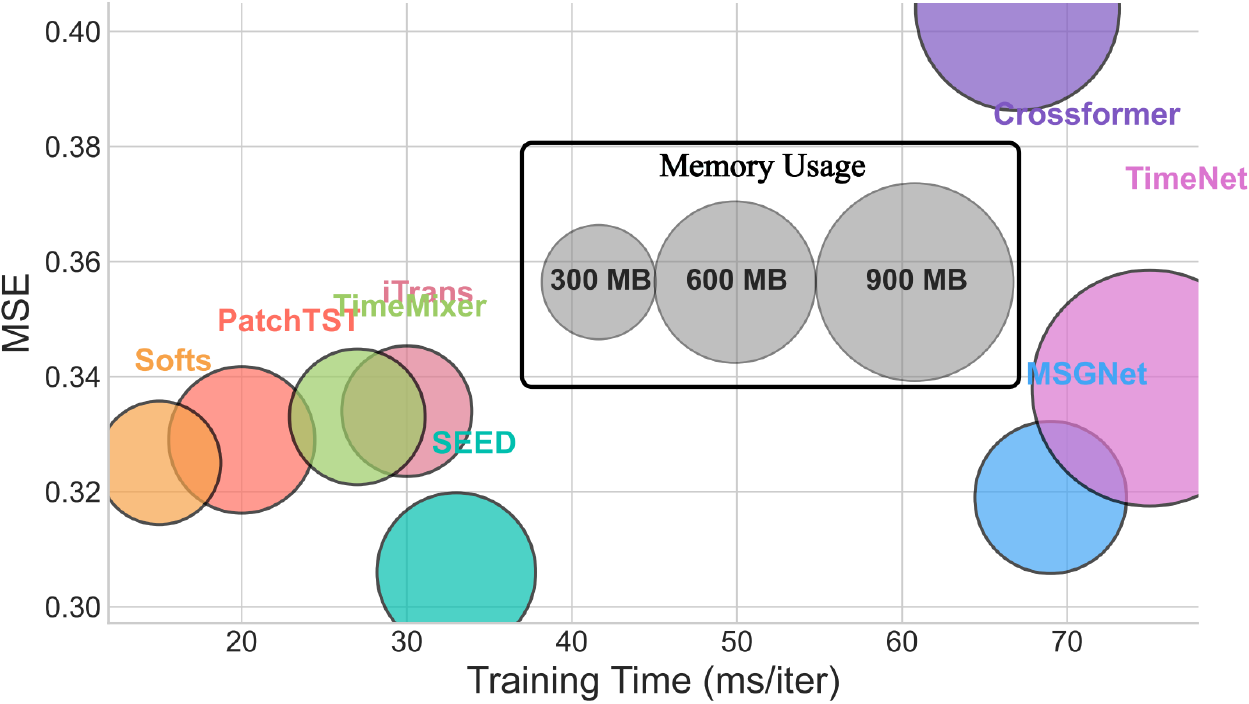}
    \caption{Performance Analysis of SEED: Assessing MSE, Training Time, and Memory Usage, evaluated on the ETTm1 Dataset with a 96-In/96-Out Setup.}
    \label{fig:ettm1_eff}
\end{figure}
We compare the SEED model against other models in terms of forecasting accuracy, memory usage, and training speed. The results, as shown in Figure \ref{fig:ettm1_eff}, indicate SEED has achieved better prediction performance while maintaining a good training speed.

\section{Conclusion}
Considering the varying complexity and internal regularity of different variables in multivariate time series, we propose SEED to adaptively evaluate and integrate spatial-temporal dependencies. Technically, the model uses spectral entropy to evaluate and measure the spatial-temporal dependencies. For complex prediction tasks, it prioritizes considering spatial dependencies, while for simpler patterns, it retains more temporal dependencies. Furthermore, our model utilizes multi-head mechanism and negative correlation to capture more intricate inter-variable relationships. Extensive experiments demonstrate that SEED consistently achieves state-of-the-art performance in both long-term and short-term forecasting tasks.

\section{Acknowledgments}
This work was supported in part by the National Natural Science Foundation of China under Grant 62376194, 625B2126; and in part by the China Scholarship Council under Grant 202406250137.

\bibliography{aaai2026}

@inproceedings{patchtst,
  title     = {A Time Series is Worth 64 Words: Long-term Forecasting with Transformers},
  author    = {Nie, Yuqi and
               H. Nguyen, Nam and
               Sinthong, Phanwadee and 
               Kalagnanam, Jayant},
  booktitle = {International Conference on Learning Representations},
  year      = {2023}
}

@inproceedings{crossformer,
  title={Crossformer: Transformer utilizing cross-dimension dependency for multivariate time series forecasting},
  author={Zhang, Yunhao and Yan, Junchi},
  booktitle={International Conference on Learning Representations},
  year={2023}
}

@article{itransformer,
  title={i{T}ransformer: Inverted Transformers Are Effective for Time Series Forecasting},
  author={Liu, Yong and Hu, Tengge and Zhang, Haoran and Wu, Haixu and Wang, Shiyu and Ma, Lintao and Long, Mingsheng},
  journal={International Conference on Learning Representations},
  year={2024},
}

@inproceedings{softs,
  title={S{OFTS}: Efficient Multivariate Time Series Forecasting with Series-Core Fusion},
  author={Han, Lu and Chen, Xu-Yang and Ye, Han-Jia and Zhan, De-Chuan},
  booktitle={Advances in Neural Information Processing Systems},
  year={2024}
}

@inproceedings{leddam,
title={Revitalizing Multivariate Time Series Forecasting: Learnable Decomposition with Inter-Series Dependencies and Intra-Series Variations Modeling},
author={Guoqi Yu and Jing Zou and Xiaowei Hu and Angelica I Aviles-Rivero and Jing Qin and Shujun Wang},
booktitle={Forty-first International Conference on Machine Learning},
year={2024},
}

@article{msgnet, 
title={MSGNet: Learning Multi-Scale Inter-series Correlations for Multivariate Time Series Forecasting}, 
journal={Proceedings of the AAAI Conference on Artificial Intelligence}, 
author={Cai, Wanlin and Liang, Yuxuan and Liu, Xianggen and Feng, Jianshuai and Wu, Yuankai}, 
year={2024}, 
}

@inproceedings{duet,
  title={Duet: Dual clustering enhanced multivariate time series forecasting},
  author={Qiu, Xiangfei and Wu, Xingjian and Lin, Yan and Guo, Chenjuan and Hu, Jilin and Yang, Bin},
  booktitle={Proceedings of the 31st ACM SIGKDD Conference on Knowledge Discovery and Data Mining V. 1},
  pages={1185--1196},
  year={2025}
}

@article{tqnet,
  title={Temporal Query Network for Efficient Multivariate Time Series Forecasting},
  author={Lin, Shengsheng and Chen, Haojun and Wu, Haijie and Qiu, Chunyun and Lin, Weiwei},
  journal={arXiv preprint arXiv:2505.12917},
  year={2025}
}

@inproceedings{bo2021beyond,
  title={Beyond low-frequency information in graph convolutional networks},
  author={Bo, Deyu and Wang, Xiao and Shi, Chuan and Shen, Huawei},
  booktitle={Proceedings of the AAAI conference on artificial intelligence},
  volume={35},
  number={5},
  pages={3950--3957},
  year={2021}
}

@article{khan2023short,
  title={Short-term traffic prediction using deep learning long short-term memory: Taxonomy, applications, challenges, and future trends},
  author={Khan, Anwar and Fouda, Mostafa M and Do, Dinh-Thuan and Almaleh, Abdulaziz and Rahman, Atiq Ur},
  journal={IEEE Access},
  volume={11},
  pages={94371--94391},
  year={2023},
  publisher={IEEE}
}

@article{baffour2019hybrid,
  title={A hybrid artificial neural network-GJR modeling approach to forecasting currency exchange rate volatility},
  author={Baffour, Alexander Amo and Feng, Jingchun and Taylor, Evans Kwesi},
  journal={Neurocomputing},
  volume={365},
  pages={285--301},
  year={2019},
  publisher={Elsevier}
}

@article{chengqing2023multi,
  title={A multi-factor driven spatiotemporal wind power prediction model based on ensemble deep graph attention reinforcement learning networks},
  author={Chengqing, Yu and Guangxi, Yan and Chengming, Yu and Yu, Zhang and Xiwei, Mi},
  journal={Energy},
  volume={263},
  pages={126034},
  year={2023},
  publisher={Elsevier}
}

@article{qiu2025comprehensive,
  title={A comprehensive survey of deep learning for multivariate time series forecasting: A channel strategy perspective},
  author={Qiu, Xiangfei and Cheng, Hanyin and Wu, Xingjian and Hu, Jilin and Guo, Chenjuan and Yang, Bin},
  journal={arXiv preprint arXiv:2502.10721},
  year={2025}
}

@article{miao2024less,
  title={Less is more: Efficient time series dataset condensation via two-fold modal matching},
  author={Miao, Hao and Liu, Ziqiao and Zhao, Yan and Guo, Chenjuan and Yang, Bin and Zheng, Kai and Jensen, Christian S},
  journal={PVLDB},
  volume={18},
  number={2},
  pages={226--238},
  year={2024}
}

@inproceedings{miao2024unified,
  title={A unified replay-based continuous learning framework for spatio-temporal prediction on streaming data},
  author={Miao, Hao and Zhao, Yan and Guo, Chenjuan and Yang, Bin and Zheng, Kai and Huang, Feiteng and Xie, Jiandong and Jensen, Christian S},
  booktitle={ICDE},
  pages={1050--1062},
  year={2024}
}

@inproceedings{liu2025timecma,
  title={Timecma: Towards llm-empowered multivariate time series forecasting via cross-modality alignment},
  author={Liu, Chenxi and Xu, Qianxiong and Miao, Hao and Yang, Sun and Zhang, Lingzheng and Long, Cheng and Li, Ziyue and Zhao, Rui},
  booktitle={AAAI},
  volume={39},
  number={18},
  pages={18780--18788},
  year={2025}
}

@inproceedings{liu2025timekd,
  title={Efficient Multivariate Time Series Forecasting via Calibrated Language Models with Privileged Knowledge Distillation},
  author={Chenxi Liu and Hao Miao and Qianxiong Xu and Shaowen Zhou and Cheng Long and Yan Zhao and Ziyue Li and Rui Zhao},
  booktitle    = {ICDE},
  year={2025}
}

@inproceedings{informer,
  title={Informer: Beyond efficient transformer for long sequence time-series forecasting},
  author={Zhou, Haoyi and Zhang, Shanghang and Peng, Jieqi and Zhang, Shuai and Li, Jianxin and Xiong, Hui and Zhang, Wancai},
  booktitle={Proceedings of the AAAI conference on artificial intelligence},
  volume={35},
  pages={11106--11115},
  year={2021}
}

@inproceedings{timesnet,
  title={Times{N}et: Temporal 2D-Variation Modeling for General Time Series Analysis},
  author={Haixu Wu and Tengge Hu and Yong Liu and Hang Zhou and Jianmin Wang and Mingsheng Long},
  booktitle={International Conference on Learning Representations},
  year={2023},
}

@inproceedings{sigir2018,
  title={Modeling long-and short-term temporal patterns with deep neural networks},
  author={Lai, Guokun and Chang, Wei-Cheng and Yang, Yiming and Liu, Hanxiao},
  booktitle={International ACM SIGIR conference on research \& development in information retrieval},
  pages={95--104},
  year={2018}
}

@article{scinet,
  title={S{CI}Net: Time series modeling and forecasting with sample convolution and interaction},
  author={Liu, Minhao and Zeng, Ailing and Chen, Muxi and Xu, Zhijian and Lai, Qiuxia and Ma, Lingna and Xu, Qiang},
  journal={Advances in Neural Information Processing Systems},
  volume={35},
  pages={5816--5828},
  year={2022}
}

@article{yi2024filternet,
  title={Filternet: Harnessing frequency filters for time series forecasting},
  author={Yi, Kun and Fei, Jingru and Zhang, Qi and He, Hui and Hao, Shufeng and Lian, Defu and Fan, Wei},
  journal={Advances in Neural Information Processing Systems},
  volume={37},
  pages={55115--55140},
  year={2024}
}

@article{shang2024ada,
  title={Ada-MSHyper: adaptive multi-scale hypergraph transformer for time series forecasting},
  author={Shang, Zongjiang and Chen, Ling and Wu, Binqing and Cui, Dongliang},
  journal={Advances in Neural Information Processing Systems},
  volume={37},
  pages={33310--33337},
  year={2024}
}

@article{chen2024similarity,
  title={From similarity to superiority: Channel clustering for time series forecasting},
  author={Chen, Jialin and Lenssen, Jan Eric and Feng, Aosong and Hu, Weihua and Fey, Matthias and Tassiulas, Leandros and Leskovec, Jure and Ying, Rex},
  journal={Advances in Neural Information Processing Systems},
  volume={37},
  pages={130635--130663},
  year={2024}
}

@article{xue2023card,
  title={Card: Channel aligned robust blend transformer for time series forecasting},
  author={Xue, Wang and Zhou, Tian and Wen, Qingsong and Gao, Jinyang and Ding, Bolin and Jin, Rong},
  journal={arXiv preprint arXiv:2305.12095},
  year={2023}
}

@inproceedings{wang2024fully,
  title={Fully-connected spatial-temporal graph for multivariate time-series data},
  author={Wang, Yucheng and Xu, Yuecong and Yang, Jianfei and Wu, Min and Li, Xiaoli and Xie, Lihua and Chen, Zhenghua},
  booktitle={Proceedings of the AAAI conference on artificial intelligence},
  volume={38},
  number={14},
  pages={15715--15724},
  year={2024}
}

@article{zhou2025crosslinear,
  title={CrossLinear: Plug-and-Play Cross-Correlation Embedding for Time Series Forecasting with Exogenous Variables},
  author={Zhou, Pengfei and Liu, Yunlong and Liang, Junli and Song, Qi and Li, Xiangyang},
  journal={arXiv preprint arXiv:2505.23116},
  year={2025}
}

@inproceedings{zhou2021informer,
  title={Informer: Beyond efficient transformer for long sequence time-series forecasting},
  author={Zhou, Haoyi and Zhang, Shanghang and Peng, Jieqi and Zhang, Shuai and Li, Jianxin and Xiong, Hui and Zhang, Wancai},
  booktitle={Proceedings of the AAAI conference on artificial intelligence},
  volume={35},
  number={12},
  pages={11106--11115},
  year={2021}
}

@article{wu2021autoformer,
  title={Autoformer: Decomposition transformers with auto-correlation for long-term series forecasting},
  author={Wu, Haixu and Xu, Jiehui and Wang, Jianmin and Long, Mingsheng},
  journal={Advances in neural information processing systems},
  volume={34},
  pages={22419--22430},
  year={2021}
}

@inproceedings{ekambaram2023tsmixer,
  title={Tsmixer: Lightweight mlp-mixer model for multivariate time series forecasting},
  author={Ekambaram, Vijay and Jati, Arindam and Nguyen, Nam and Sinthong, Phanwadee and Kalagnanam, Jayant},
  booktitle={Proceedings of the 29th ACM SIGKDD conference on knowledge discovery and data mining},
  pages={459--469},
  year={2023}
}

@article{yi2023fouriergnn,
  title={FourierGNN: Rethinking multivariate time series forecasting from a pure graph perspective},
  author={Yi, Kun and Zhang, Qi and Fan, Wei and He, Hui and Hu, Liang and Wang, Pengyang and An, Ning and Cao, Longbing and Niu, Zhendong},
  journal={Advances in neural information processing systems},
  volume={36},
  pages={69638--69660},
  year={2023}
}

@article{zhao2023multiple,
  title={Multiple time series forecasting with dynamic graph modeling},
  author={Zhao, Kai and Guo, Chenjuan and Cheng, Yunyao and Han, Peng and Zhang, Miao and Yang, Bin},
  journal={Proceedings of the VLDB Endowment},
  volume={17},
  number={4},
  pages={753--765},
  year={2023},
  publisher={VLDB Endowment}
}

@article{sun2021solar,
  title={Solar Wind Speed Prediction With Two-Dimensional Attention Mechanism},
  author={Sun, Yanru and Xie, Zongxia and Chen, Yanhong and Huang, Xin and Hu, Qinghua},
  journal={Space Weather},
  volume={19},
  number={7},
  pages={e2020SW002707},
  year={2021},
  publisher={Wiley Online Library}
}

@article{sun2022accurate,
  title={Accurate solar wind speed prediction with multimodality information},
  author={Sun, Yanru and Xie, Zongxia and Chen, Yanhong and Hu, Qinghua},
  journal={Space: Science \& Technology},
  year={2022},
  publisher={AAAS}
}

@article{sun2025ppgf,
  title={PPGF: Probability Pattern-Guided Time Series Forecasting},
  author={Sun, Yanru and Xie, Zongxia and Xing, Haoyu and Yu, Hualong and Hu, Qinghua},
  journal={IEEE Transactions on Neural Networks and Learning Systems},
  year={2025},
  publisher={IEEE}
}

@inproceedings{sun2025hierarchical,
  title={Hierarchical classification auxiliary network for time series forecasting},
  author={Sun, Yanru and Xie, Zongxia and Chen, Dongyue and Eldele, Emadeldeen and Hu, Qinghua},
  booktitle={Proceedings of the AAAI Conference on Artificial Intelligence},
  volume={39},
  number={19},
  pages={20743--20751},
  year={2025}
}

@inproceedings{sun2025learning,
  title={Learning Pattern-Specific Experts for Time Series Forecasting Under Patch-level Distribution Shift},
  author={Yanru Sun and Zongxia Xie and Emadeldeen Eldele and Dongyue Chen and Qinghua Hu and Min Wu},
  booktitle={The Thirty-ninth Annual Conference on Neural Information Processing Systems},
  year={2025},
  url={https://openreview.net/forum?id=CtoIG9Iwas}
}

@article{ma2024survey,
  title={A survey on time-series pre-trained models},
  author={Ma, Qianli and Liu, Zhen and Zheng, Zhenjing and Huang, Ziyang and Zhu, Siying and Yu, Zhongzhong and Kwok, James T},
  journal={IEEE Transactions on Knowledge and Data Engineering},
  year={2024},
  publisher={IEEE}
}

@inproceedings{
liu2025learning,
title={Learning Soft Sparse Shapes for Efficient Time-Series Classification},
author={Zhen Liu and Yicheng Luo and Boyuan Li and Emadeldeen Eldele and Min Wu and Qianli Ma},
booktitle={Forty-second International Conference on Machine Learning},
year={2025},
url={https://openreview.net/forum?id=B9DOjtj9xK}
}

@inproceedings{
luo2025hipatch,
title={Hi-Patch: Hierarchical Patch {GNN} for Irregular Multivariate Time Series},
author={Yicheng Luo and Bowen Zhang and Zhen Liu and Qianli Ma},
booktitle={Forty-second International Conference on Machine Learning},
year={2025},
url={https://openreview.net/forum?id=nBgQ66iEUu}
}

@article{wang2023drift,
  title={Drift doesn't matter: Dynamic decomposition with diffusion reconstruction for unstable multivariate time series anomaly detection},
  author={Wang, Chengsen and Zhuang, Zirui and Qi, Qi and Wang, Jingyu and Wang, Xingyu and Sun, Haifeng and Liao, Jianxin},
  journal={Advances in neural information processing systems},
  volume={36},
  pages={10758--10774},
  year={2023}
}

@inproceedings{wang2025chattime,
  title={Chattime: A unified multimodal time series foundation model bridging numerical and textual data},
  author={Wang, Chengsen and Qi, Qi and Wang, Jingyu and Sun, Haifeng and Zhuang, Zirui and Wu, Jinming and Zhang, Lei and Liao, Jianxin},
  booktitle={Proceedings of the AAAI Conference on Artificial Intelligence},
  volume={39},
  number={12},
  pages={12694--12702},
  year={2025}
}

@inproceedings{wang2024interdependency,
  title={Interdependency matters: graph alignment for multivariate time series anomaly detection},
  author={Wang, Yuanyi and Sun, Haifeng and Wang, Chengsen and Zhu, Mengde and Wang, Jingyu and Tang, Wei and Qi, Qi and Zhuang, Zirui and Liao, Jianxin},
  booktitle={2024 IEEE International Conference on Data Mining (ICDM)},
  pages={869--874},
  year={2024},
  organization={IEEE}
}

@article{qiu2024tfb,
  title={TFB: Towards Comprehensive and Fair Benchmarking of Time Series Forecasting Methods},
  author={Qiu, Xiangfei and Hu, Jilin and Zhou, Lekui and Wu, Xingjian and Du, Junyang and Zhang, Buang and Guo, Chenjuan and Zhou, Aoying and Jensen, Christian S and Sheng, Zhenli and others},
  journal={Proceedings of the VLDB Endowment},
  volume={17},
  number={9},
  pages={2363--2377},
  year={2024},
  publisher={VLDB Endowment}
}

@inproceedings{
wu2025catch,
title={{CATCH}: Channel-Aware Multivariate Time Series Anomaly Detection via Frequency Patching},
author={Xingjian Wu and Xiangfei Qiu and Zhengyu Li and Yihang Wang and Jilin Hu and Chenjuan Guo and Hui Xiong and Bin Yang},
booktitle={The Thirteenth International Conference on Learning Representations},
year={2025},
url={https://openreview.net/forum?id=m08aK3xxdJ}
}

\clearpage

\newpage
\appendix
\section*{\centering{APPENDIX}}

\setcounter{figure}{0}
\renewcommand{\thefigure}{A\arabic{figure}}   
\setcounter{table}{0}
\renewcommand{\thetable}{A\arabic{table}}     
\setcounter{equation}{0}
\renewcommand{\theequation}{A\arabic{equation}} 

\section{ Spectral Entropy and Autocorrelation Function }

This appendix discusses the relationship between spectral entropy and autocorrelation function.

\subsection{ Autocorrelation Function:}
Let \( x(t) \) be a wide-sense stationary real-valued stochastic process with zero mean. The \emph{autocorrelation function}(ACF) \( R(\tau) \) is defined as:
\begin{equation}
R(\tau) = \mathbb{E}[x(t) \cdot x(t + \tau)]
\end{equation}
which describes the temporal dependency structure of the process. To investigate its frequency characteristics, we examine the \emph{power spectral density} (PSD), denoted by \( S(f) \), which characterizes how the signal's power is distributed over frequency.

\subsection{ Derivation of the Frequency Representation:}

We begin with the finite-time Fourier transform of \( x(t) \) over the interval \( [-T, T] \):
\begin{equation}
X_T(f) = \int_{-T}^{T} x(t) e^{-2\pi i f t} \, dt
\end{equation}
The empirical power spectrum is then defined as:
\begin{equation}
S_T(f) = \frac{1}{2T} |X_T(f)|^2 = \frac{1}{2T} X_T(f) \overline{X_T(f)}
\end{equation}
Expanding the square magnitude yields:
\begin{equation}
S_T(f) = \frac{1}{2T} \int_{-T}^{T} \int_{-T}^{T} x(t) x(s) e^{-2\pi i f (t-s)} \, dt \, ds
\end{equation}
Let us change variables: \( \tau = t-s \), \( u = s \), so \( t = u + \tau \). Then the integral becomes:
\begin{equation}
\begin{aligned}
&S_T(f) = \\
&\int_{-2T}^{2T} \left( \frac{1}{2T} \int_{\max(-T, -T+\tau)}^{\min(T, T+\tau)} x(u+\tau) x(u) \, du \right) e^{-2\pi i f \tau} \, d\tau
\end{aligned}
\end{equation}

Taking expectation and applying the stationarity assumption \( \mathbb{E}[x(u+\tau)x(u)] = R(\tau) \), we get:
\begin{equation}
\mathbb{E}[S_T(f)] \approx \int_{-\infty}^{\infty} R(\tau) e^{-2\pi i f \tau} \, d\tau
\end{equation}
Therefore, in the limit as \( T \to \infty \), we obtain:
\begin{equation}
S(f) = \int_{-\infty}^{\infty} R(\tau) e^{-2\pi i f \tau} \, d\tau
\end{equation}
That is, the power spectral density \( S(f) \) is the \emph{Fourier transform of the autocorrelation function} \( R(\tau) \).

\subsection{ Spectral Entropy:}

The power spectral density \( S(f) \) can be normalized to define a probability density function over frequencies:
\begin{equation}
P(f) = \frac{S(f)}{\int_{-\infty}^{\infty} S(f) \, df}, \quad \text{with } \int P(f) \, df = 1
\end{equation}
The \emph{spectral entropy} is defined analogously to Shannon entropy:
\begin{equation}
H_s =-\int P(f) \log P(f) \, df
\end{equation}

Spectral entropy quantifies the concentration of spectral energy:
\begin{itemize}
  \item Low spectral entropy implies energy is concentrated in specific frequencies, corresponding to strong periodic or autocorrelated structure.
  \item High spectral entropy indicates a flat spectrum, as in white noise or unstructured signals.
\end{itemize}

\subsection{Visualization of Spectral Entropy and ACF:}
\begin{figure}
    \centering
    \includegraphics[width=1\linewidth]{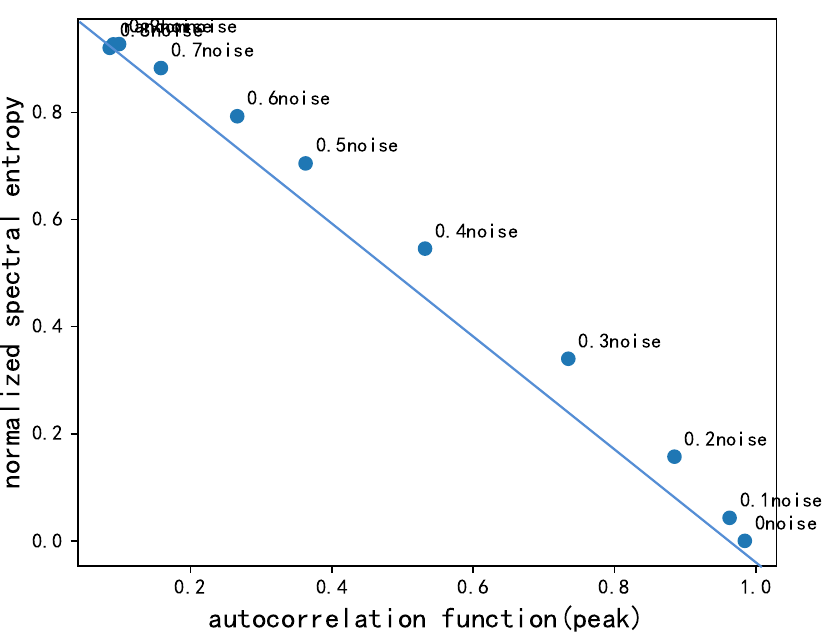}
    \caption{The relationship between the autocorrelation function and the normalized spectral entropy. The decimals before the "noise" label represent the proportion of noise.}
    \label{fig:acf_vs_se}
\end{figure}

\begin{figure*}[t!]
    \centering
    \includegraphics[width=0.8\linewidth]{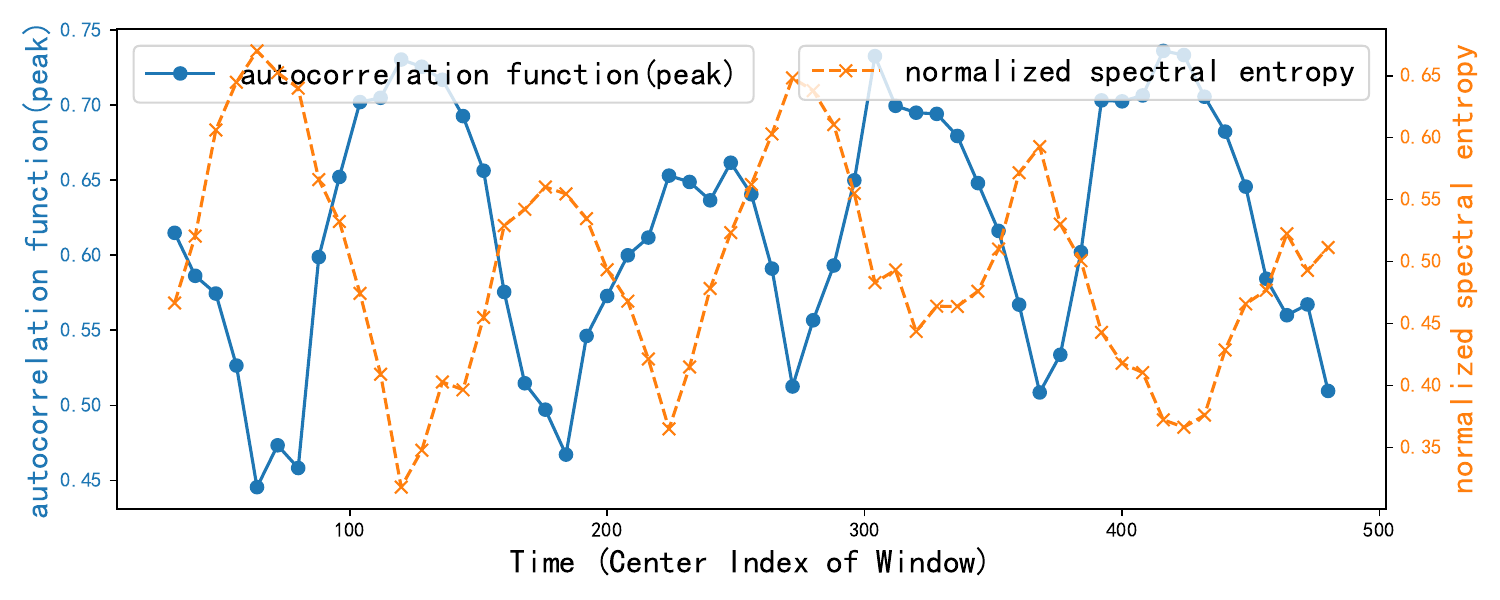}
    \caption{On the constructed signal, the autocorrelation function (peak) and the normalized spectral entropy are calculated using a moving window to investigate their relationship.}
    \label{fig:acf_se_win}
\end{figure*}

In order to more clearly and directly understand the relationship between the two, we artificially constructed a set of data with different noise proportions. 
\begin{equation}
x_t = (1-\alpha)f_{\text{per}}(t) + \alpha\varepsilon_t, 
\varepsilon_t \sim \mathcal{N}(0,1) \label{eq:signal_noise_set_gene}
\end{equation}

where $, t=1,2,\dots,T$, $f_{\text{per}}(t)$ denotes a periodic signal, and $\alpha$ controls the noise, $\varepsilon_t$ denotes the noise.
We calculated the maximum value in the autocorrelation function and the normalized spectral entropy respectively, and visualized them as Figure \ref{fig:acf_vs_se}. In Figure A1, the decimal number before the "noise" label is exactly the "alpha" in the aforementioned formula \ref{eq:signal_noise_set_gene}.
Furthermore, we calculated the spectral entropy and peak values of the autocorrelation function for each individual signal generated using the aforementioned formula, and plotted Figure \ref{fig:acf_se_win}.

As shown in the figure \ref{fig:acf_vs_se} and \ref{fig:acf_se_win}, the autocorrelation function and spectral entropy exhibit a strong negative correlation, and both the autocorrelation function and the spectral entropy can effectively measure the general range of the noise.

\subsection{Comparison Spectral Entropy and ACF:}
Although there is a certain degree of similarity between Spectral Entropy and ACF, ACF only considers the correlation at a specific lag order, identifying more of a linear dependence. As shown in Formulas A6 and A7, the ACF has an advantage in reflecting the main peak in the frequency domain. However, when multiple peaks are considered, the fixed lag limit restricts its ability to perceive multiple peaks. In the real world, the dependencies of time series are often complex and nonlinear signals. In contrast, spectral entropy can identify more abundant internal regularity patterns (such as multi-periodicity, etc.), and can quantify the complexity, disorder, and randomness of time series. Moreover, in the identification of linear dependence patterns, the above visualization verifies the substitutability of spectral entropy for it. Table \ref{tab:results_se_acf} demonstrates the superiority of spectral entropy over the autocorrelation function.

\begin{table}[t]
\centering
\begin{tabular}{c|c|c|c|c|c}
\toprule
\multicolumn{2}{c}{\multirow{2}{*}{\scalebox{1.1}{Models}}} & \multicolumn{2}{c|}{SEED (SE)} & \multicolumn{2}{c}{SEED(ACF)} \\
\cmidrule{3-6}
\multicolumn{2}{c}{Metric}  & {MSE} & {MAE} & {MSE} & {MAE} \\
\midrule
\multirow{5}{*}{ETTm1} & 96 & \textbf{0.306} & \textbf{0.351} & 0.317 & 0.359 \\
& 192 & \textbf{0.350} & \textbf{0.378} & 0.363 & 0.384 \\
& 336 & \textbf{0.381} & \textbf{0.400} & 0.397 & 0.406 \\
& 720 & \textbf{0.441} & \textbf{0.435} & 0.461 & 0.442 \\
\cmidrule(lr){2-6}
& avg & \textbf{0.369} & \textbf{0.391} & 0.385 & 0.398 \\
\midrule
\multirow{5}{*}{ETTm2} & 96 & \textbf{0.169} & \textbf{0.255} & 0.172 & 0.257 \\
& 192 & \textbf{0.234} & \textbf{0.297} & 0.236 & 0.298 \\
& 336 & \textbf{0.297} & \textbf{0.337} & 0.298 & 0.337 \\
& 720 & \textbf{0.393} & \textbf{0.395} & 0.398 & 0.396 \\
\cmidrule(lr){2-6}
& avg & \textbf{0.273} & \textbf{0.321} & 0.276 & 0.322 \\
\midrule
\multirow{5}{*}{ETTh1} & 96 & \textbf{0.370} & \textbf{0.392} & 0.379 & 0.394 \\
& 192 & \textbf{0.422} & 0.425 & 0.428 & \textbf{0.422} \\
& 336 & \textbf{0.443} & \textbf{0.440} & 0.468 & 0.447 \\
& 720 & \textbf{0.438} & \textbf{0.453} & 0.463 & 0.464 \\
\cmidrule(lr){2-6}
& avg & \textbf{0.418} & \textbf{0.428} & 0.435 & 0.432 \\
\midrule
\multirow{5}{*}{ETTh2} & 96 & \textbf{0.282} & \textbf{0.336} & 0.286 & 0.336 \\
& 192 & \textbf{0.362} & \textbf{0.388} & 0.368 & 0.390 \\
& 336 & \textbf{0.401} & \textbf{0.421} & 0.405 & 0.421 \\
& 720 & \textbf{0.418} & \textbf{0.439} & 0.453 & 0.459 \\
\cmidrule(lr){2-6}
& avg & \textbf{0.365} & \textbf{0.396} & 0.378 & 0.402 \\
\midrule
\multirow{5}{*}{Weather} & 96 & \textbf{0.151} & \textbf{0.198} & 0.158 & 0.204 \\
& 192 & \textbf{0.202} & \textbf{0.247} & 0.205 & 0.248 \\
& 336 & \textbf{0.260} & 0.292 & 0.266 & \textbf{0.291} \\
& 720 & \textbf{0.343} & \textbf{0.346} & 0.350 & 0.346 \\
\cmidrule(lr){2-6}
& avg & \textbf{0.239} & \textbf{0.271} & 0.245 & 0.272 \\
\midrule
\multirow{5}{*}{ECL} & 96 & \textbf{0.133} & \textbf{0.230} & 0.140 & 0.235 \\
& 192 & \textbf{0.151} & \textbf{0.245} & 0.168 & 0.258 \\
& 336 & \textbf{0.160} & \textbf{0.257} & 0.324 & 0.391 \\
& 720 & \textbf{0.179} & \textbf{0.279} & 0.214 & 0.300 \\
\cmidrule(lr){2-6}
& avg & \textbf{0.156} & \textbf{0.253} & 0.212 & 0.296 \\
\midrule
\multirow{5}{*}{Traffic} & 96 & \textbf{0.370} & \textbf{0.247} & 0.373 & 0.249 \\
& 192 & \textbf{0.392} & \textbf{0.258} & 0.393 & 0.260 \\
& 336 & \textbf{0.412} & \textbf{0.269} & 0.414 & 0.272 \\
& 720 & \textbf{0.444} & \textbf{0.289} & 0.447 & 0.292 \\
\cmidrule(lr){2-6}
& avg & \textbf{0.404} & \textbf{0.266} & 0.407 & 0.268 \\
\midrule
\multirow{5}{*}{Solar-Energy} & 96 & \textbf{0.189} & \textbf{0.227} & \textbf{0.189} & \textbf{0.227} \\
& 192 & \textbf{0.213} & \textbf{0.249} & 0.225 & 0.257 \\
& 336 & \textbf{0.234} & \textbf{0.265} & 0.237 & 0.269 \\
& 720 & \textbf{0.239} & \textbf{0.270} & 0.244 & 0.275 \\
\cmidrule(lr){2-6}
& avg & \textbf{0.219} & \textbf{0.253} & 0.224 & 0.257 \\
\bottomrule
\end{tabular}
\caption{Comparison of MSE and MAE for SEED (SE) and (ACF) across various datasets. The input sequence length $L$ is set to $96$ for all variants. \textit{Avg} are  averaged across four different forecasting horizon: $T \in \{96, 192, 336, 720\}$. The best results are highlighted in \textbf{bold}.}
\label{tab:results_se_acf}
\end{table}
\section{Implementation Details}
\subsection{Dataset Descriptions}\label{app:dataset}
We conduct extensive experiments on eight widely-used time series datasets for long-term forecasting and PEMS datasets for short-term forecasting. We report the statistics in \ref{tab:dataset}. Detailed descriptions of these datasets are as follows:

\begin{enumerate}
\item [(1)] \textbf{ETT} (Electricity Transformer Temperature) dataset \citep{informer} encompasses temperature and power load data from electricity transformers in two regions of China, spanning from 2016 to 2018. This dataset has two granularity levels: ETTh (hourly) and ETTm (15 minutes).
\item [(2)] \textbf{Weather} dataset \citep{timesnet} captures 21 distinct meteorological indicators in Germany, meticulously recorded at 10-minute intervals throughout 2020. Key indicators in this dataset include air temperature, visibility, among others, offering a comprehensive view of the weather dynamics.
\item [(3)] \textbf{Electricity} dataset \citep{timesnet} features hourly electricity consumption records in kilowatt-hours (kWh) for 321 clients. Sourced from the UCL Machine Learning Repository, this dataset covers the period from 2012 to 2014, providing valuable insights into consumer electricity usage patterns.
\item [(4)] \textbf{Traffic} dataset \citep{timesnet} includes data on hourly road occupancy rates, gathered by 862 detectors across the freeways of the San Francisco Bay area. This dataset, covering the years 2015 to 2016, offers a detailed snapshot of traffic flow and congestion.
\item [(5)] \textbf{Solar-Energy} dataset \citep{sigir2018} contains solar power production data recorded every 10 minutes throughout 2006 from 137 photovoltaic (PV) plants in Alabama. 
\item [(6)] \textbf{PEMS} dataset \citep{scinet} comprises four public traffic network datasets (PEMS03, PEMS04, PEMS07, and PEMS08), constructed from the Caltrans Performance Measurement System (PeMS) across four districts in California. The data is aggregated into 5-minute intervals, resulting in 12 data points per hour and 288 data points per day.
\end{enumerate}

\renewcommand{\arraystretch}{1.0}
\begin{table*}[htb]
  \centering
   \resizebox{\textwidth}{!}{
  \begin{threeparttable}
  \renewcommand{\multirowsetup}{\centering}
  \setlength{\tabcolsep}{3pt}
  \begin{tabular}{c|l|c|c|c|c}
    \toprule
    Tasks & Dataset & Dim & Prediction Length & Split & Frequency \\ 
    \toprule
     & ETTm1 & $7$ & \scalebox{0.8}{$\{96, 192, 336, 720\}$} & $6:2:2$  & $15$ min \\ 
    \cmidrule{2-6}
    & ETTm2 & $7$ & \scalebox{0.8}{$\{96, 192, 336, 720\}$} & $6:2:2$  & $15$ min \\ 
    \cmidrule{2-6}
     & ETTh1 & $7$ & \scalebox{0.8}{$\{96, 192, 336, 720\}$} & $6:2:2$ & $15$ min \\ 
    \cmidrule{2-6}
     Long-term & ETTh2 & $7$ & \scalebox{0.8}{$\{96, 192, 336, 720\}$} & $6:2:2$ & $15$ min \\ 
    \cmidrule{2-6}
    Forecasting & Weather & $21$ & \scalebox{0.8}{$\{96, 192, 336, 720\}$} & $7:1:2$  & $10$ min \\ 
    \cmidrule{2-6}
     & ECL & $321$ & \scalebox{0.8}{$\{96, 192, 336, 720\}$} & $7:1:2$ & $1$ hour \\ 
    \cmidrule{2-6}
     & Traffic & $862$ & \scalebox{0.8}{$\{96, 192, 336, 720\}$} & $7:1:2$ & $1$ hour \\ 
    \cmidrule{2-6}
    & Solar & $137$  & \scalebox{0.8}{$\{96, 192, 336, 720\}$}  & $7:1:2$ & $10$ min \\ 
    \midrule

    & PEMS03 & $358$ & ${12, 24, 48}$ & $3:1:1$ & $5$ min \\ 
    \cmidrule{2-6}
    Short-term & PEMS04 & $307$ & ${12, 24, 48}$ & $3:1:1$ & $5$ min \\ 
    \cmidrule{2-6}
    Forecasting & PEMS07 & $883$ & ${12, 24, 48}$ & $3:1:1$ & $5$ min \\ 
    \cmidrule{2-6}
    & PEMS08 & $170$ & ${12, 24, 48}$ & $3:1:1$ & $5$ min \\ 
    
    \bottomrule
    \end{tabular}
  \end{threeparttable}
}
\caption{
Detailed dataset descriptions. \textit{Dim} denotes the variate number of each dataset. \textit{Split} represents the proportion of data points allocated to the training, validation, and testing sets, respectively. \textit{Prediction Length} denotes the future time points to be predicted.\textit{Frequency} denotes the sampling interval of time points.}
\label{tab:dataset}
\end{table*}

\subsection{Metric Details}
Regarding metrics, we utilize the mean square error (MSE) and mean absolute error (MAE) for long-term forecasting. The calculations of these metrics are:
\begin{equation}
MSE=\frac{1}{T}\sum_{0}^{T}(\hat{y}_{i}-y_i)^2\ ,\ 
MAE=\frac{1}{T}\sum_{0}^{T}|\hat{y}_{i}-y_i| \nonumber
\end{equation}

\section{Full Results}
\subsection{Long-term Forecasting}

Table \ref{tab::app_full_long_result} presents the full results for long-term forecasting, the look-back horizon is $L=96$ and the forecasting horizon $T\in\{96, 192, 336, 720\}$.

\subsection{Short-term Forecasting}

Table \ref{tab::app_full_short_result} presents the full results for long-term forecasting, the look-back horizon is $L=96$ and the forecasting horizon $T\in\{12,24,48\}$. Across all four PEMS datasets, SEED consistently demonstrated the best performance. On most datasets, SEED maintains competitive performance. However, the reason why SEED lags behind TQNet on the Solar-Energy dataset might be that TQNet employs global periodic relationships, which enables it to aware longer periodicities beyond the window.

\subsection{Ablation Study}

We present the full results of the ablation studies discussed in the main text in table \ref{tab::app_full_ablation_result}. To validate the effectiveness of \textbf{SEED}, we construct the following model variants by removing or replacing key components:

\begin{itemize}
    \item \textbf{w/o-TAttn}: Removes the Temporal Attention (TAttn) module to evaluate the importance of temporal dependency modeling.
    
    \item \textbf{w/o-CSE}: Removes the Context Spatial Extractor (CSE) module to assess the contribution of spatial context modeling.

    \item \textbf{re-S1}: Replaces the Signed Graph Constructor (SGC) with a softmax-based graph, which ignores negative correlations, to examine the impact of signed edge modeling.

    \item \textbf{re-S2}: Applies softmax to the magnitude of correlations while decoupling the sign, forming a signed-softmax graph. This tests whether softmax-based signed graphs preserve correlation polarity.

    \item \textbf{re-F1}: Replaces the spectral entropy-based fusion weights with a learnable scalar parameter to analyze the benefit of dynamic uncertainty-aware fusion.

    \item \textbf{re-F2}: Swaps the outputs of the TAttn and CSE modules before fusion to validate the directional roles of each branch.

    \item \textbf{re-F3}: Concatenates the outputs from both branches and feeds them into a linear layer to predict adaptive fusion weights, serving as a learning-based fusion baseline.

    \item \textbf{re-C1}: Modifies the CSE module to model spatial dependencies only among variables at the same time step, thereby ignoring cross-temporal interactions.

    \item \textbf{re-C2}: Uses a fully-connected graph across all temporal patches in the CSE module, which increases expressiveness but ignores localized inductive biases.
\end{itemize}

\subsection{Hyper-Parameter Sensitivity}
\begin{figure*}
    \centering
    \includegraphics[width=1\linewidth]{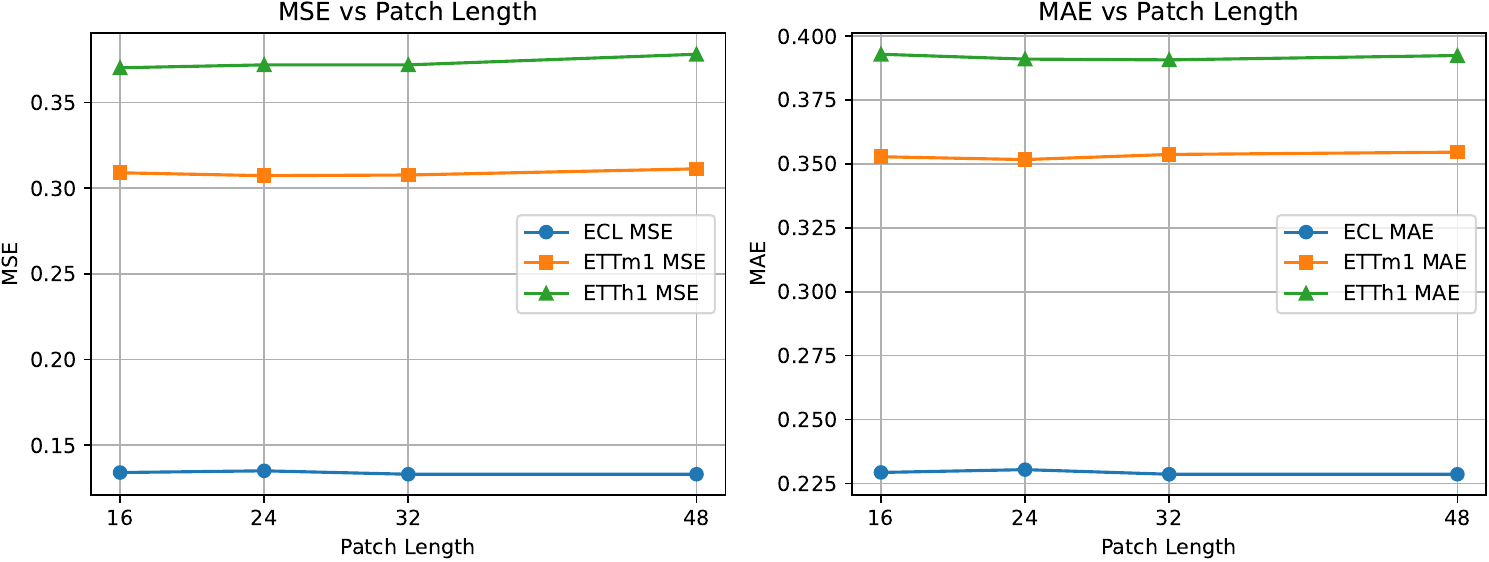}
    \caption{Hyperparameter sensitivity with the length of the patch. The results are recorded with the lookback window length $L$ = 96 and the forecast window length $T$ = 96.}
    \label{fig:para}
\end{figure*}

We evaluate the hyperparameter sensitivity of SEED with the length of the patch, and the results are shown in Figure \ref{fig:para}. As can be seen from the figure \ref{fig:para}, our model is not sensitive to the length of the patch.
\subsection{Visualization}

\subsubsection{Visualization of spatial-temporal features}
\begin{figure}[htb]
    \centering
    \includegraphics[width=0.9\linewidth]{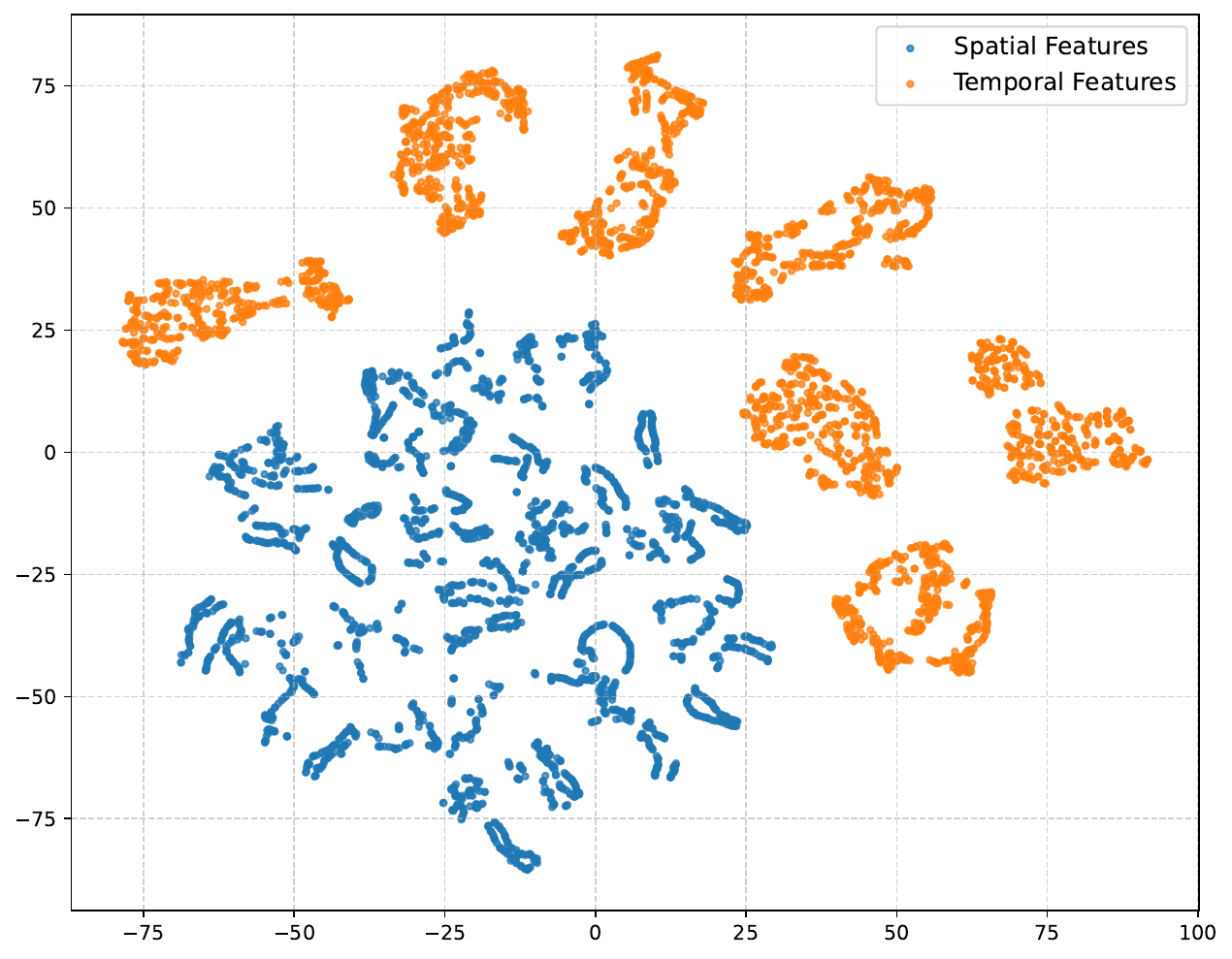}
    \caption{T-SNE visualization of spatial-temporal features on the ETTm1 dataset.}
    \label{fig:ettm1_tsne}
\end{figure}

\begin{figure}[htb]
    \centering
    \includegraphics[width=0.9\linewidth]{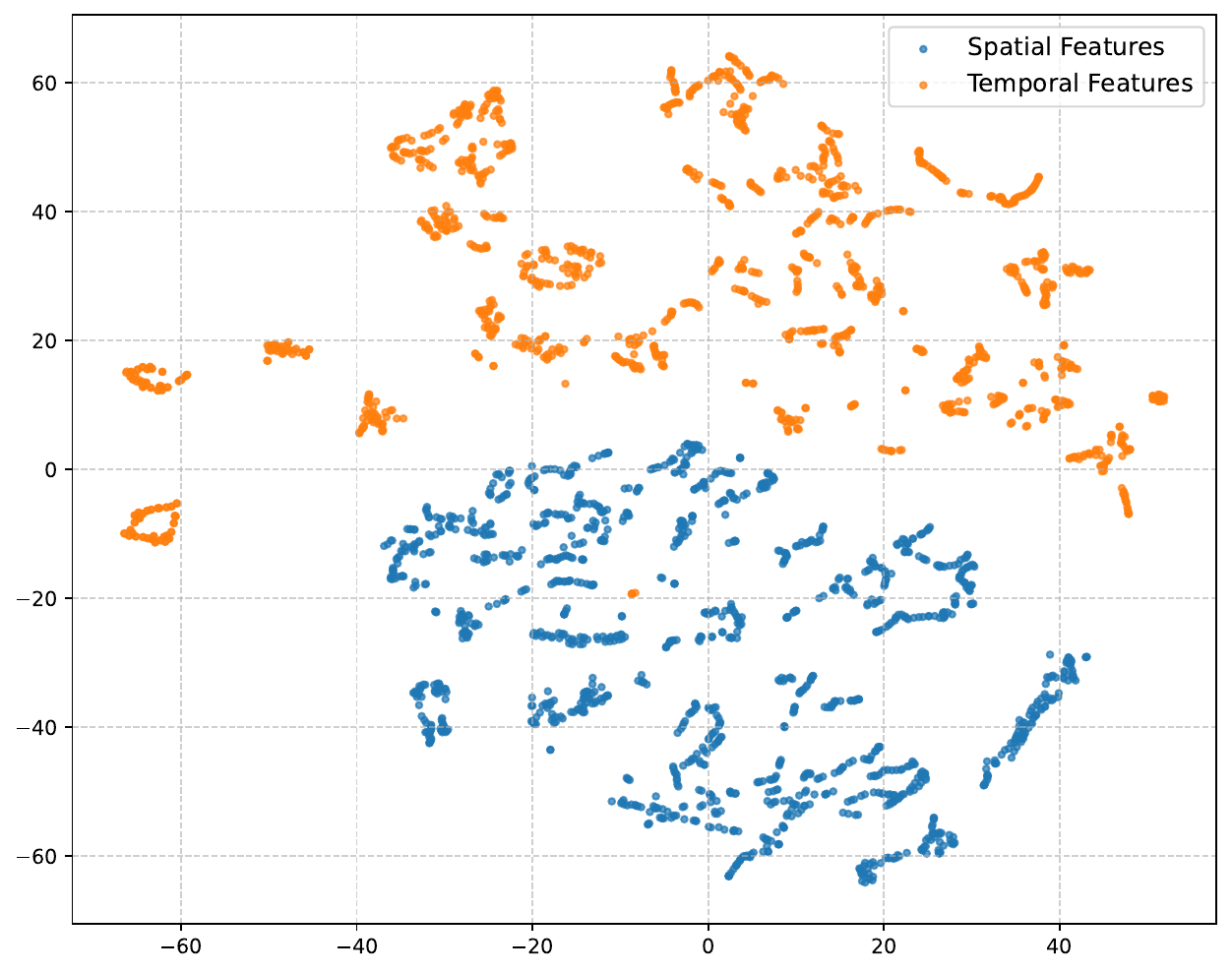}
    \caption{T-SNE visualization of spatial-temporal features on the Weather dataset.}
    \label{fig:weather_tsne}
\end{figure}

\begin{figure}[htb]
    \centering
    \includegraphics[width=0.9\linewidth]{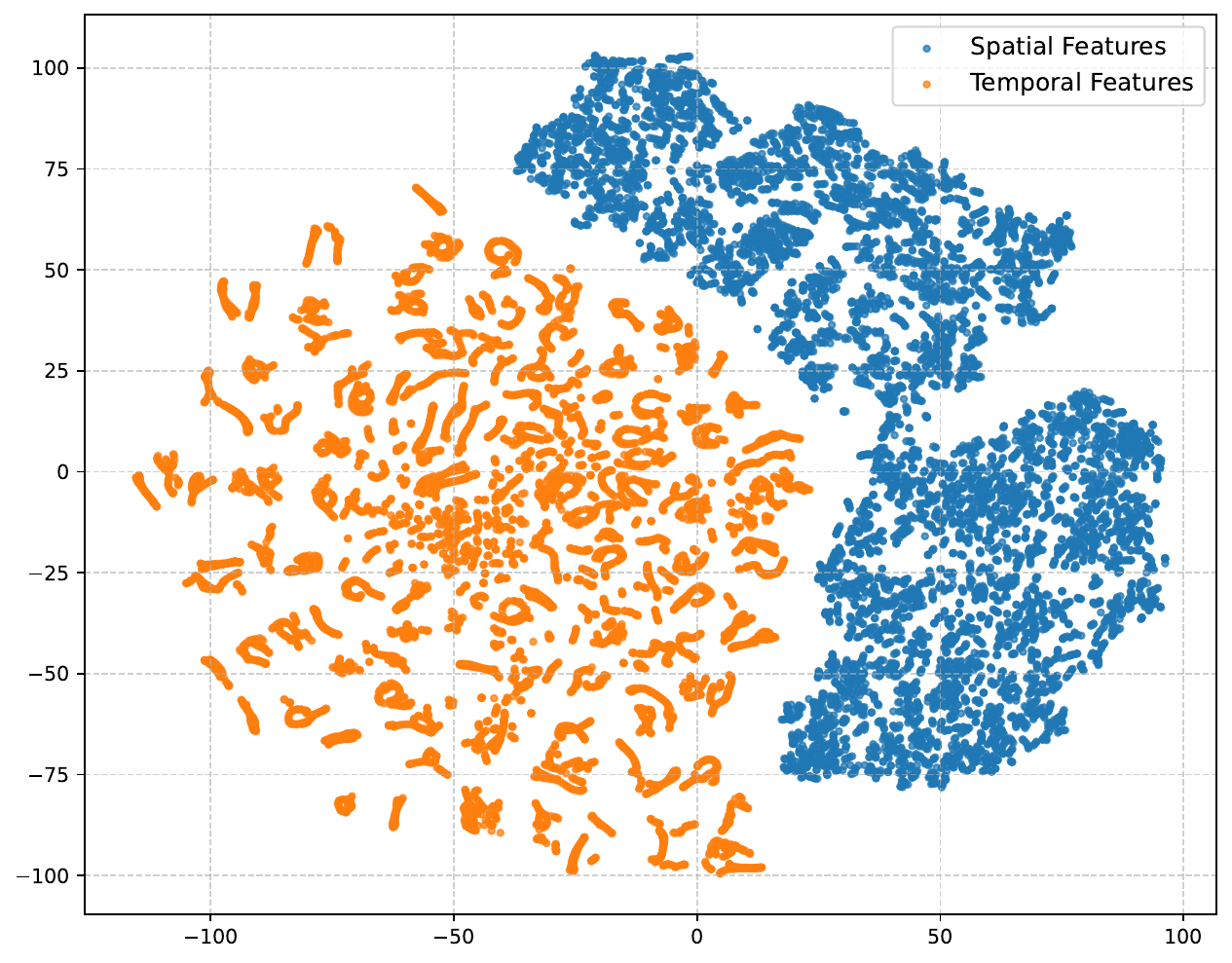}
    \caption{T-SNE visualization of spatial-temporal features on the Solar dataset.}
    \label{fig:solar_tsne}
\end{figure}

\begin{figure}[htb]
    \centering
    \includegraphics[width=0.9\linewidth]{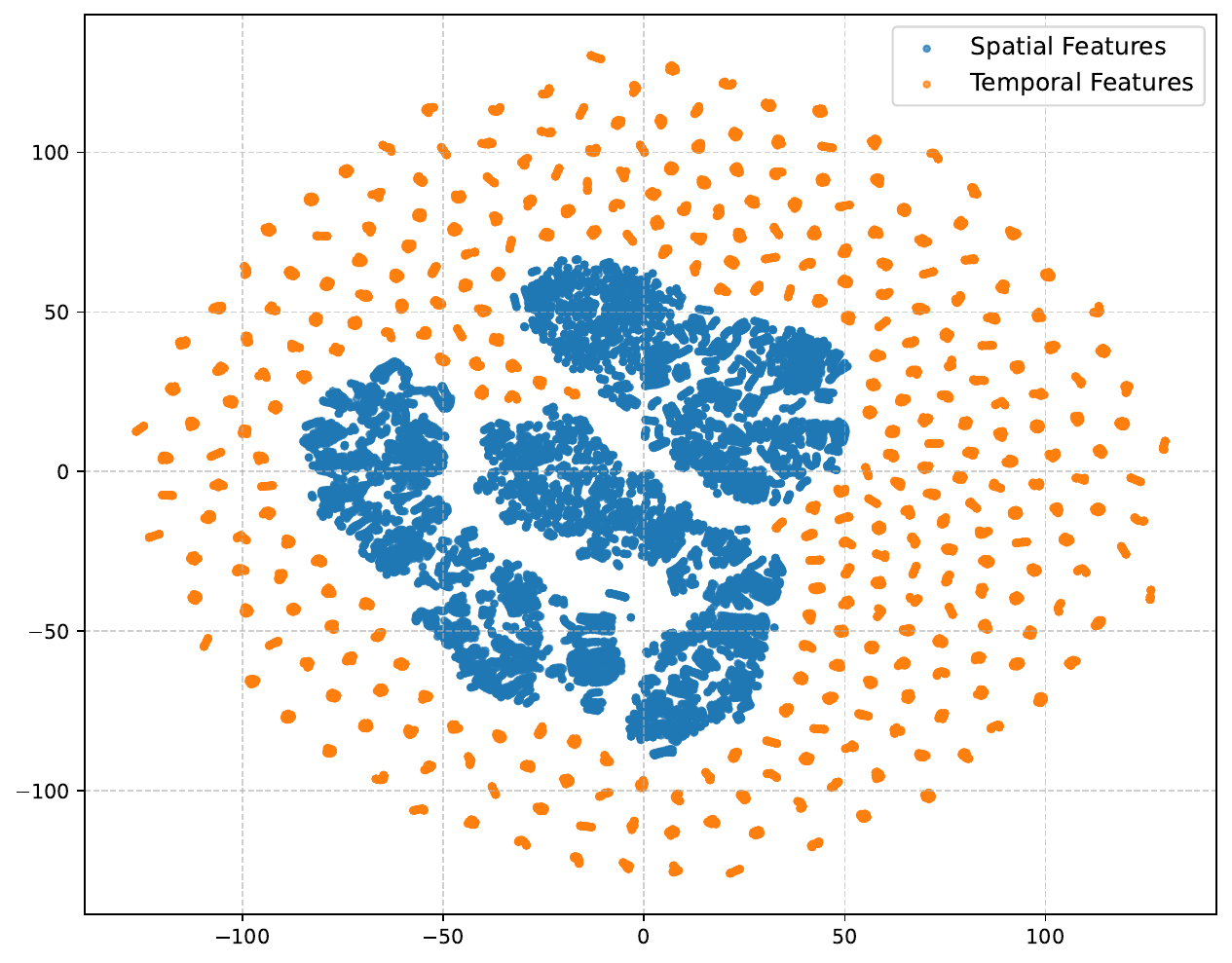}
    \caption{T-SNE visualization of spatial-temporal features on the ECL dataset.}
    \label{fig:ecl_tsne}
\end{figure}

To explore the relationship between spatial and temporal features, we perform t-SNE visualization on the spatiotemporal features of our model. We present the visualization results of spatiotemporal features for one batch, with ETTm1 shown in Figure \ref{fig:ettm1_tsne}, Weather in Figure \ref{fig:weather_tsne}, Solar in Figure \ref{fig:solar_tsne} and Electricity in Figure \ref{fig:ecl_tsne}. Here, orange represents temporal features and blue represents spatial features. It can be observed that temporal features, which rely only on themselves, appear relatively independent and scattered, while spatial features, which depend on each other, exhibit a more dense and concentrated pattern.
In the ETTm1 dataset, the variables are relatively independent and display periodic variations. As illustrated in Figure \ref{fig:ettm1_tsne}, the orange cluster contains seven clusters, which precisely correspond to the seven variables in ETTm1, demonstrating that our model successfully retains the intrinsic temporal dynamics of each variable. This property is also partially reflected in the Solar and ECL datasets. The Weather dataset lacks a clear periodic pattern and exhibits more complex variable variation.



\subsubsection{Visualization of Prediction Results}
\begin{figure*}[hb]
    \centering
    \includegraphics[width=1\linewidth]{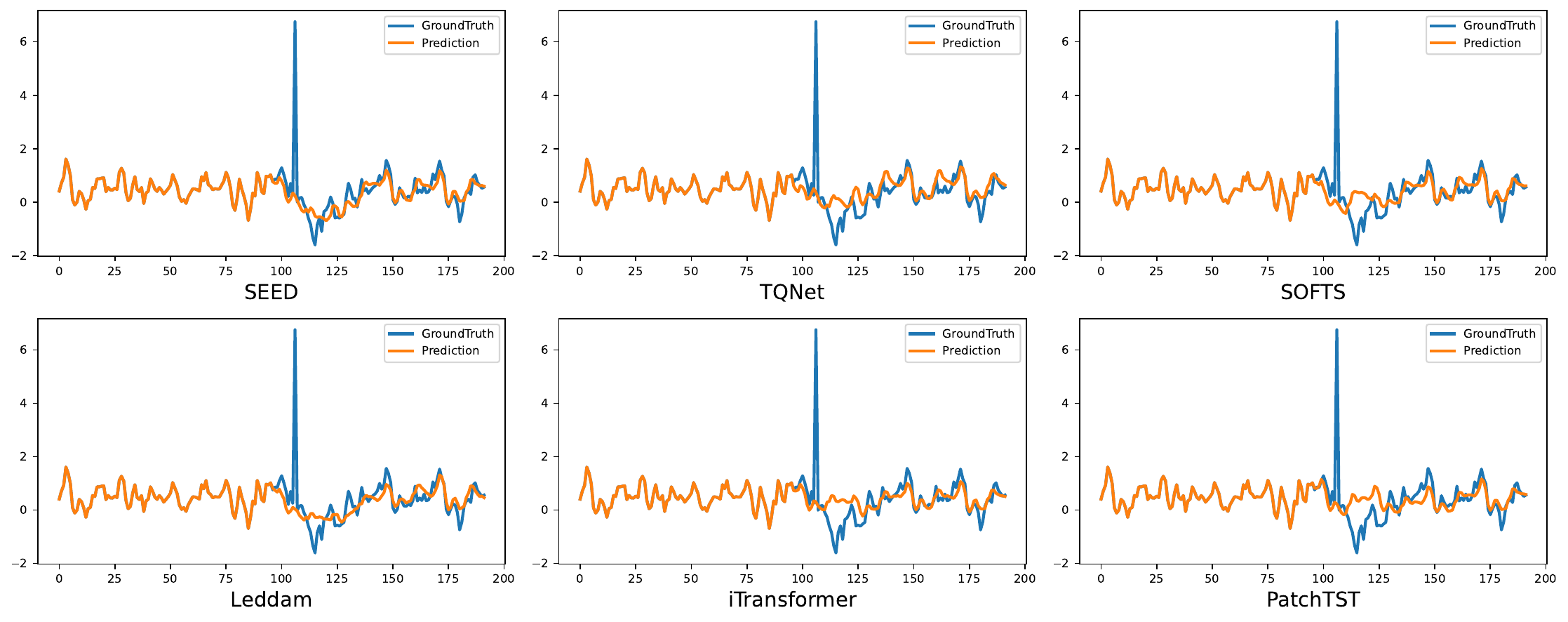}
    \caption{Visualization of predictions from different models on the ECL dataset.}
    \label{fig:ecl_case}
\end{figure*}

\begin{figure*}[htb]
    \centering
    \includegraphics[width=1\linewidth]{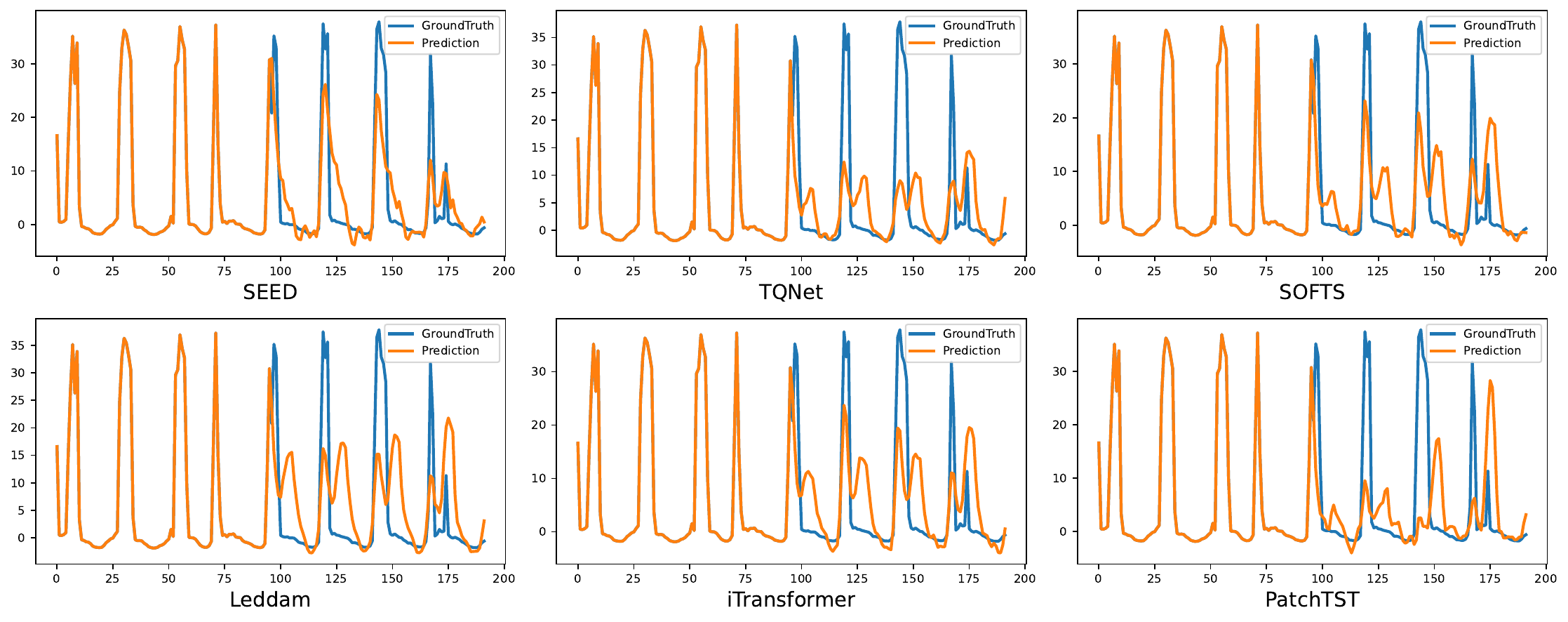}
    \caption{Visualization of predictions from different models on the Traffic dataset.}
    \label{fig:traffic_case}
\end{figure*}

In order to facilitate a clear comparison between different models, we present supplementary prediction examples  for two representative datasets, as Electricity in Figure \ref{fig:ecl_case}, Traffic in Figure \ref{fig:traffic_case}, respectively. The look-back horizon L is 96 and the forecasting horizon T is 96. Among the different models, SEED delivers the most  accurate predictions of future series variations and demonstrates superior performance. As shown in the Figure \ref{fig:traffic_case}, SEED accurately predicted a more reasonable main peak based on historical data, and captured the double peak in the last section (which might be caused by the influence of other variables). This indicates that our model can well balance the influence of its own characteristics and external other variables.

\begin{table*}[t!]
\centering
\begin{tabular}{c|c|cc|cc|cc|cc|cc|cc}

\toprule
\multicolumn{2}{c}{\multirow{2}{*}{\scalebox{1.1}{Models}}} & \multicolumn{2}{c}{SEED} & \multicolumn{2}{c}{TQNet} & \multicolumn{2}{c}{DUET} & \multicolumn{2}{c}{iTransformer} & \multicolumn{2}{c}{MSGNet} & \multicolumn{2}{c}{SOFTS} \\
\multicolumn{2}{c}{} & \multicolumn{2}{c}{\scalebox{0.8}{(\textbf{Ours})}} & \multicolumn{2}{c}{\scalebox{0.8}{\citeyearpar{tqnet}}} & \multicolumn{2}{c}{\scalebox{0.8}{\citeyearpar{duet}}} & \multicolumn{2}{c}{\scalebox{0.8}{\citeyearpar{itransformer}}} & \multicolumn{2}{c}{\scalebox{0.8}{\citeyearpar{msgnet}}} & \multicolumn{2}{c}{\scalebox{0.8}{\citeyearpar{softs}}} \\

\cmidrule(lr){3-4} \cmidrule(lr){5-6} \cmidrule(lr){7-8} \cmidrule(lr){9-10} \cmidrule(lr){11-12} \cmidrule(lr){13-14}

\multicolumn{2}{c}{Metric} & \scalebox{0.85}{MSE} & \scalebox{0.85}{MAE} & \scalebox{0.85}{MSE} & \scalebox{0.85}{MAE} & \scalebox{0.85}{MSE} & \scalebox{0.85}{MAE} & \scalebox{0.85}{MSE} & \scalebox{0.85}{MAE} & \scalebox{0.85}{MSE} & \scalebox{0.85}{MAE} & \scalebox{0.85}{MSE} & \scalebox{0.85}{MAE} \\

\toprule

\multirow{5}{*}{\rotatebox[origin=c]{90}{ETTm1}}
& 96 & \textbf{0.306} & \textbf{0.351} & \underline{0.311} & \underline{0.353} & 0.324 & 0.354 & 0.334 & 0.368 & 0.319 & 0.366 & 0.325 & 0.361 \\
& 192 & \textbf{0.350} & \textbf{0.378} & \underline{0.356} & \textbf{0.378} & 0.369 & \underline{0.379} & 0.377 & 0.391 & 0.376 & 0.397 & 0.375 & 0.389 \\
& 336 & \textbf{0.381} & \textbf{0.400} & \underline{0.390} & \underline{0.401} & 0.404 & 0.402 & 0.426 & 0.420 & 0.417 & 0.422 & 0.405 & 0.412 \\
& 720 & \textbf{0.441} & \textbf{0.435} & \underline{0.452} & 0.440 & 0.463 & \underline{0.437} & 0.491 & 0.459 & 0.481 & 0.458 & 0.466 & 0.447 \\
\cmidrule(lr){2-14}
& \emph{Avg.} & \textbf{0.369} & \textbf{0.391} & \underline{0.377} & \underline{0.393} & 0.390 & \underline{0.393} & 0.407 & 0.410 & 0.398 & 0.411 & 0.393 & 0.403 \\

\midrule
\multirow{5}{*}{\rotatebox[origin=c]{90}{ETTm2}}
& 96 & \textbf{0.169} & \textbf{0.255} & \underline{0.173} & \underline{0.256} & 0.174 & \textbf{0.255} & 0.180 & 0.264 & 0.177 & 0.262 & 0.180 & 0.261 \\
& 192 & \textbf{0.234} & \textbf{0.297} & \underline{0.238} & \underline{0.298} & 0.243 & 0.302 & 0.250 & 0.309 & 0.247 & 0.307 & 0.246 & 0.306 \\
& 336 & \textbf{0.297} & \textbf{0.337} & \underline{0.301} & \underline{0.340} & 0.304 & 0.341 & 0.311 & 0.348 & 0.312 & 0.346 & 0.319 & 0.352 \\
& 720 & \textbf{0.393} & \textbf{0.395} & \underline{0.397} & \underline{0.396} & 0.399 & 0.397 & 0.412 & 0.407 & 0.414 & 0.403 & 0.405 & 0.401 \\
\cmidrule(lr){2-14}
& \emph{Avg.} & \textbf{0.273} & \textbf{0.321} & \underline{0.277} & \underline{0.323} & 0.280 & 0.324 & 0.288 & 0.332 & 0.288 & 0.330 & 0.287 & 0.330 \\

\midrule
\multirow{5}{*}{\rotatebox[origin=c]{90}{ETTh1}}
& 96 & \textbf{0.370} & \textbf{0.392} & \underline{0.371} & \underline{0.393} & 0.377 & \underline{0.393} & 0.386 & 0.405 & 0.390 & 0.411 & 0.381 & 0.399 \\
& 192 & \textbf{0.422} & \textbf{0.425} & \underline{0.428} & 0.426 & 0.429 & 0.425 & 0.441 & 0.436 & 0.442 & 0.442 & 0.435 & \underline{0.431} \\
& 336 & \textbf{0.443} & \textbf{0.440} & \underline{0.476} & 0.456 & 0.471 & \underline{0.446} & 0.487 & 0.458 & 0.480 & 0.468 & 0.480 & 0.452 \\
& 720 & \textbf{0.438} & \textbf{0.453} & \underline{0.487} & \underline{0.470} & 0.496 & 0.480 & 0.503 & 0.491 & 0.494 & 0.488 & 0.499 & 0.488 \\
\cmidrule(lr){2-14}
& \emph{Avg.} & \textbf{0.418} & \textbf{0.428} & \underline{0.441} & \underline{0.434} & 0.443 & 0.436 & 0.454 & 0.447 & 0.452 & 0.452 & 0.449 & 0.442 \\

\midrule
\multirow{5}{*}{\rotatebox[origin=c]{90}{ETTh2}}
& 96 & \textbf{0.282} & \textbf{0.336} & 0.295 & \underline{0.343} & \underline{0.296} & 0.345 & 0.297 & 0.349 & 0.328 & 0.371 & 0.297 & 0.347 \\
& 192 & \textbf{0.362} & \textbf{0.388} & \underline{0.367} & 0.393 & 0.368 & \underline{0.389} & 0.380 & 0.400 & 0.402 & 0.414 & 0.373 & 0.394 \\
& 336 & \textbf{0.401} & \textbf{0.421} & 0.417 & 0.427 & \underline{0.411} & \underline{0.422} & 0.428 & 0.432 & 0.435 & 0.443 & 0.410 & 0.426 \\
& 720 & 0.418 & 0.439 & 0.433 & 0.446 & \underline{0.412} & 0.434 & 0.427 & 0.445 & 0.417 & 0.441 & \textbf{0.411} & \textbf{0.433} \\
\cmidrule(lr){2-14}
& \emph{Avg.} & \textbf{0.365} & \textbf{0.396} & 0.378 & 0.402 & \underline{0.372} & \underline{0.397} & 0.383 & 0.407 & 0.396 & 0.417 & 0.385 & 0.408 \\

\midrule
\multirow{5}{*}{\rotatebox[origin=c]{90}{Weather}}
& 96 & \textbf{0.151} & \textbf{0.198} & \underline{0.157} & \underline{0.200} & 0.163 & 0.202 & 0.174 & 0.214 & 0.163 & 0.212 & 0.166 & 0.208 \\
& 192 & \textbf{0.202} & 0.247 & \underline{0.206} & \textbf{0.245} & 0.218 & 0.252 & 0.221 & 0.254 & 0.212 & 0.254 & 0.217 & \underline{0.253} \\
& 336 & \textbf{0.260} & 0.292 & \underline{0.262} & \textbf{0.287} & 0.274 & 0.294 & 0.278 & \underline{0.296} & 0.272 & 0.299 & 0.282 & 0.300 \\
& 720 & 0.343 & 0.346 & \underline{0.344} & \textbf{0.342} & 0.349 & \underline{0.343} & 0.358 & 0.349 & 0.350 & 0.348 & 0.356 & 0.351 \\
\cmidrule(lr){2-14}
& \emph{Avg.} & \textbf{0.239} & \underline{0.270} & \underline{0.242} & \textbf{0.269} & 0.251 & 0.273 & 0.258 & 0.279 & 0.249 & 0.278 & 0.255 & 0.278 \\

\midrule
\multirow{5}{*}{\rotatebox[origin=c]{90}{Electricity}}
& 96 & \textbf{0.133} & \textbf{0.230} & \underline{0.134} & \textbf{0.229} & 0.145 & 0.233 & 0.148 & 0.240 & 0.165 & 0.274 & 0.143 & 0.233 \\
& 192 & \textbf{0.151} & \textbf{0.245} & \underline{0.154} & \underline{0.247} & 0.163 & 0.248 & 0.162 & 0.253 & 0.184 & 0.292 & 0.158 & 0.248 \\
& 336 & \textbf{0.160} & \textbf{0.257} & \underline{0.169} & 0.264 & 0.175 & \underline{0.262} & 0.178 & 0.269 & 0.195 & 0.302 & 0.178 & 0.269 \\
& 720 & \textbf{0.179} & \textbf{0.279} & \underline{0.201} & 0.294 & 0.204 & \underline{0.291} & 0.225 & 0.317 & 0.231 & 0.332 & 0.218 & 0.305 \\
\cmidrule(lr){2-14}
& \emph{Avg.} & \textbf{0.156} & \textbf{0.253} & \underline{0.164} & 0.259 & 0.172 & \underline{0.258} & 0.178 & 0.270 & 0.194 & 0.300 & 0.174 & 0.264 \\

\midrule
\multirow{5}{*}{\rotatebox[origin=c]{90}{Traffic}}
& 96 & \textbf{0.370} & \textbf{0.247} & 0.413 & 0.261 & 0.407 & 0.252 & \underline{0.395} & 0.268 & 0.605 & 0.344 & 0.376 & \underline{0.251} \\
& 192 & \textbf{0.392} & \textbf{0.258} & 0.432 & 0.271 & 0.431 & \underline{0.262} & 0.417 & 0.276 & 0.613 & 0.359 & \underline{0.398} & 0.261 \\
& 336 & \textbf{0.412} & \textbf{0.269} & 0.450 & \underline{0.277} & 0.456 & \textbf{0.269} & 0.433 & 0.283 & 0.642 & 0.376 & \underline{0.415} & \textbf{0.269} \\
& 720 & \textbf{0.444} & 0.289 & 0.486 & 0.295 & 0.509 & 0.292 & 0.467 & 0.302 & 0.702 & 0.401 & \underline{0.447} & \textbf{0.287} \\
\cmidrule(lr){2-14}
& \emph{Avg.} & \textbf{0.404} & \textbf{0.266} & 0.445 & 0.276 & 0.451 & 0.269 & 0.428 & 0.282 & 0.641 & 0.370 & \underline{0.409} & \underline{0.267} \\

\midrule
\multirow{5}{*}{\rotatebox[origin=c]{90}{Solar-Energy}}
& 96 & \underline{0.189} & \underline{0.227} & \textbf{0.173} & 0.233 & 0.200 & \textbf{0.207} & 0.203 & 0.237 & 0.208 & 0.243 & 0.200 & 0.230 \\
& 192 & \underline{0.213} & \underline{0.249} & \textbf{0.199} & 0.257 & 0.228 & \textbf{0.233} & 0.233 & 0.261 & 0.258 & 0.281 & 0.229 & 0.253 \\
& 336 & 0.234 & 0.265 & \textbf{0.211} & \underline{0.263} & 0.262 & \textbf{0.244} & 0.248 & 0.273 & 0.293 & 0.311 & \underline{0.243} & 0.269 \\
& 720 & \underline{0.239} & \textbf{0.270} & \textbf{0.209} & \textbf{0.270} & 0.258 & \underline{0.249} & 0.249 & 0.275 & 0.290 & 0.315 & 0.245 & 0.272 \\
\cmidrule(lr){2-14}
& \emph{Avg.} & \underline{0.219} & \underline{0.253} & \textbf{0.198} & 0.256 & 0.237 & \textbf{0.233} & 0.233 & 0.262 & 0.262 & 0.288 & 0.229 & 0.256 \\

\bottomrule
\end{tabular}
\caption{Full results of long-term forecasting. The input sequence
length $L$ is set to $96$ for all baselines. \textit{Avg} are  averaged across four different forecasting horizon: $T \in \{96, 192, 336, 720\}$.  The best results are highlighted in \textbf{bold}, while the second-best results are \underline{underlined}.}
\label{tab::app_full_long_result}
\end{table*}

\begin{table*}[t!]
\centering
\begin{tabular}{c|c|cc|cc|cc|cc|cc|cc}

\toprule
\multicolumn{2}{c}{\multirow{2}{*}{\scalebox{1.1}{Models}}} & \multicolumn{2}{c}{SEED} & \multicolumn{2}{c}{iTransformer} & \multicolumn{2}{c}{Leddam} & \multicolumn{2}{c}{SOFTS} & \multicolumn{2}{c}{PatchTST} & \multicolumn{2}{c}{Crossformer} \\
\multicolumn{2}{c}{} & \multicolumn{2}{c}{\scalebox{0.8}{(\textbf{Ours})}} & \multicolumn{2}{c}{\scalebox{0.8}{(2024b)}} & \multicolumn{2}{c}{\scalebox{0.8}{(2024)}} & \multicolumn{2}{c}{\scalebox{0.8}{(2024a)}} & \multicolumn{2}{c}{\scalebox{0.8}{(2023)}} & \multicolumn{2}{c}{\scalebox{0.8}{(2023)}} \\

\cmidrule(lr){3-4} \cmidrule(lr){5-6} \cmidrule(lr){7-8} \cmidrule(lr){9-10} \cmidrule(lr){11-12} \cmidrule(lr){13-14}

\multicolumn{2}{c}{Metric} & \scalebox{0.85}{MSE} & \scalebox{0.85}{MAE} & \scalebox{0.85}{MSE} & \scalebox{0.85}{MAE} & \scalebox{0.85}{MSE} & \scalebox{0.85}{MAE} & \scalebox{0.85}{MSE} & \scalebox{0.85}{MAE} & \scalebox{0.85}{MSE} & \scalebox{0.85}{MAE} & \scalebox{0.85}{MSE} & \scalebox{0.85}{MAE} \\

\toprule

\multirow{4}{*}{\rotatebox[origin=c]{90}{PEMS03}}
& 12 & \textbf{0.063} & \textbf{0.164} & 0.071 & 0.174 & 0.068 & 0.174 & \underline{0.064} & \underline{0.165} & 0.099 & 0.216 & 0.090 & 0.203 \\
& 24 & \textbf{0.079} & \textbf{0.183} & 0.093 & 0.201 & 0.094 & 0.202 & \underline{0.083} & \underline{0.188} & 0.142 & 0.259 & 0.121 & 0.240 \\
& 48 & \textbf{0.112} & \textbf{0.219} & 0.125 & 0.236 & 0.140 & 0.254 & \underline{0.114} & \underline{0.223} & 0.211 & 0.319 & 0.202 & 0.317 \\
\cmidrule(lr){2-14}
& \emph{Avg.} & \textbf{0.085} & \textbf{0.189} & 0.096 & 0.204 & 0.101 & 0.210 & \underline{0.087} & \underline{0.192} & 0.151 & 0.265 & 0.138 & 0.253 \\

\midrule
\multirow{4}{*}{\rotatebox[origin=c]{90}{PEMS04}}
& 12 & \textbf{0.066} & \textbf{0.165} & 0.078 & 0.183 & \underline{0.076} & 0.182 & 0.074 & \underline{0.176} & 0.105 & 0.224 & 0.098 & 0.218 \\
& 24 & \textbf{0.078} & \textbf{0.180} & 0.095 & 0.207 & 0.097 & 0.209 & \underline{0.088} & \underline{0.194} & 0.153 & 0.257 & 0.131 & 0.256 \\
& 48 & \textbf{0.097} & \textbf{0.203} & 0.120 & 0.233 & 0.132 & 0.249 & \underline{0.110} & \underline{0.219} & 0.229 & 0.339 & 0.205 & 0.326 \\
\cmidrule(lr){2-14}
& \emph{Avg.} & \textbf{0.080} & \textbf{0.183} & 0.098 & 0.207 & 0.102 & 0.213 & \underline{0.091} & \underline{0.196} & 0.162 & 0.273 & 0.145 & 0.267 \\

\midrule
\multirow{4}{*}{\rotatebox[origin=c]{90}{PEMS07}}
& 12 & \textbf{0.054} & \textbf{0.147} & 0.067 & 0.165 & 0.066 & 0.164 & \underline{0.057} & \underline{0.152} & 0.095 & 0.207 & 0.094 & 0.200 \\
& 24 & \textbf{0.066} & \textbf{0.163} & 0.088 & 0.190 & 0.079 & 0.185 & \underline{0.073} & \underline{0.173} & 0.150 & 0.262 & 0.139 & 0.247 \\
& 48 & \textbf{0.084} & \textbf{0.186} & 0.110 & 0.215 & 0.115 & 0.228 & \underline{0.096} & \underline{0.195} & 0.253 & 0.340 & 0.311 & 0.369 \\
\cmidrule(lr){2-14}
& \emph{Avg.} & \textbf{0.068} & \textbf{0.165} & 0.088 & 0.190 & 0.087 & 0.192 & \underline{0.075} & \underline{0.173} & 0.166 & 0.270 & 0.181 & 0.272 \\

\midrule
\multirow{4}{*}{\rotatebox[origin=c]{90}{PEMS08}}
& 12 & \textbf{0.062} & \textbf{0.160} & 0.079 & 0.182 & \underline{0.070} & \underline{0.173} & 0.074 & 0.171 & 0.168 & 0.232 & 0.165 & 0.214 \\
& 24 & \textbf{0.077} & \textbf{0.179} & 0.115 & 0.219 & \underline{0.091} & \underline{0.200} & 0.104 & 0.201 & 0.224 & 0.281 & 0.215 & 0.260 \\
& 48 & \textbf{0.102} & \textbf{0.209} & 0.186 & \underline{0.235} & 0.145 & 0.261 & 0.164 & 0.253 & 0.321 & 0.354 & 0.315 & 0.335 \\
\cmidrule(lr){2-14}
& \emph{Avg.} & \textbf{0.080} & \textbf{0.183} & 0.127 & 0.212 & \underline{0.102} & 0.211 & 0.114 & 0.208 & 0.238 & 0.289 & 0.232 & 0.270 \\

\bottomrule
\end{tabular}
\caption{Full results of short-term forecasting. The input sequence
length is set to 96 for all baselines. \textit{Avg} are  averaged across three different forecasting horizon: $T \in \{12, 24, 48\}$. The best results are highlighted in \textbf{bold}, while the second-best results are \underline{underlined}.}
\label{tab::app_full_short_result}
\end{table*}

\begin{table*}[t!]
\centering
\scalebox{0.68}{
\begin{tabular}{c|c|cc|cc|cc|cc|cc|cc|cc|cc|cc|cc}
\toprule
\multicolumn{2}{c}{\multirow{2}{*}{\scalebox{1.1}{Variants}}} & \multicolumn{2}{c}{SEED} & \multicolumn{2}{c}{w/o TAttn} & \multicolumn{2}{c}{w/o CSE} & \multicolumn{2}{c}{re-S1} & \multicolumn{2}{c}{re-S2} & \multicolumn{2}{c}{re-F1} & \multicolumn{2}{c}{re-F2} & \multicolumn{2}{c}{re-F3} & \multicolumn{2}{c}{re-C1} & \multicolumn{2}{c}{re-C2} \\
\cmidrule(lr){3-4} \cmidrule(lr){5-6} \cmidrule(lr){7-8} \cmidrule(lr){9-10} \cmidrule(lr){11-12} \cmidrule(lr){13-14} \cmidrule(lr){15-16} \cmidrule(lr){17-18} \cmidrule(lr){19-20} \cmidrule(lr){21-22}
\multicolumn{2}{c}{Metric} & \scalebox{0.85}{MSE} & \scalebox{0.85}{MAE} & \scalebox{0.85}{MSE} & \scalebox{0.85}{MAE} & \scalebox{0.85}{MSE} & \scalebox{0.85}{MAE} & \scalebox{0.85}{MSE} & \scalebox{0.85}{MAE} & \scalebox{0.85}{MSE} & \scalebox{0.85}{MAE} & \scalebox{0.85}{MSE} & \scalebox{0.85}{MAE} & \scalebox{0.85}{MSE} & \scalebox{0.85}{MAE} & \scalebox{0.85}{MSE} & \scalebox{0.85}{MAE} & \scalebox{0.85}{MSE} & \scalebox{0.85}{MAE} & \scalebox{0.85}{MSE} & \scalebox{0.85}{MAE} \\
\toprule
\multirow{5}{*}{\rotatebox[origin=c]{90}{ETTm1}}
& 96 & \textbf{0.306} & \textbf{0.351} & 0.313 & 0.354 & 0.316 & \underline{0.356} & 0.310 & \underline{0.355} & 0.311 & \underline{0.356} & \underline{0.316} & \underline{0.355} & 0.318 & 0.360 & 0.313 & 0.354 & \underline{0.308} & \underline{0.354} & \underline{0.309} & \underline{0.354} \\
& 192 & \textbf{0.350} & \underline{0.378} & \underline{0.355} & \textbf{0.377} & 0.356 & \underline{0.379} & \underline{0.354} & 0.381 & \underline{0.354} & 0.382 & \underline{0.356} & \underline{0.379} & 0.361 & 0.383 & \underline{0.355} & \textbf{0.377} & 0.351 & \underline{0.378} & 0.356 & 0.381 \\
& 336 & \underline{0.381} & \textbf{0.400} & 0.388 & \underline{0.401} & \underline{0.384} & \textbf{0.400} & \underline{0.384} & 0.404 & 0.386 & 0.405 & \underline{0.383} & \underline{0.399} & 0.393 & 0.406 & \underline{0.382} & \underline{0.399} & \underline{0.384} & 0.403 & \textbf{0.381} & \underline{0.400} \\
& 720 & \textbf{0.441} & \underline{0.435} & 0.448 & \underline{0.436} & 0.445 & \underline{0.435} & 0.445 & 0.439 & 0.445 & 0.439 & 0.445 & \underline{0.435} & 0.448 & 0.439 & 0.445 & \underline{0.436} & \underline{0.439} & \underline{0.435} & 0.442 & \textbf{0.435} \\
\cmidrule(lr){2-22}
& \emph{Avg.} & \textbf{0.369} & \textbf{0.391} & 0.376 & \underline{0.392} & \underline{0.375} & 0.393 & \underline{0.373} & 0.395 & 0.374 & 0.396 & \underline{0.375} & \underline{0.392} & 0.380 & 0.397 & \underline{0.374} & \underline{0.392} & \underline{0.371} & \underline{0.393} & 0.372 & \underline{0.392} \\
\midrule
\multirow{5}{*}{\rotatebox[origin=c]{90}{ETTm2}}
& 96 & \underline{0.169} & \textbf{0.255} & \textbf{0.167} & \textbf{0.253} & 0.171 & \underline{0.257} & \textbf{0.167} & \textbf{0.253} & 0.171 & 0.256 & 0.172 & 0.258 & 0.170 & \underline{0.257} & 0.170 & 0.256 & \underline{0.169} & \textbf{0.255} & 0.172 & 0.258 \\
& 192 & \underline{0.234} & \textbf{0.297} & \textbf{0.233} & \textbf{0.296} & 0.235 & \underline{0.299} & \textbf{0.233} & \textbf{0.296} & 0.238 & 0.300 & 0.236 & 0.300 & 0.238 & 0.300 & 0.235 & 0.300 & 0.239 & 0.303 & 0.237 & 0.302 \\
& 336 & \underline{0.297} & \textbf{0.337} & \textbf{0.294} & \underline{0.335} & \underline{0.295} & 0.340 & \underline{0.295} & \underline{0.335} & 0.305 & 0.341 & 0.296 & 0.341 & 0.300 & 0.339 & \underline{0.297} & 0.340 & 0.300 & 0.339 & 0.302 & 0.341 \\
& 720 & \underline{0.393} & \underline{0.395} & \textbf{0.393} & \textbf{0.394} & \textbf{0.393} & 0.399 & \textbf{0.393} & \textbf{0.394} & 0.420 & 0.406 & \textbf{0.393} & 0.400 & 0.399 & \underline{0.399} & \underline{0.392} & 0.398 & 0.399 & \underline{0.395} & 0.397 & 0.396 \\
\cmidrule(lr){2-22}
& \emph{Avg.} & \underline{0.273} & \textbf{0.321} & \textbf{0.272} & \textbf{0.320} & \underline{0.274} & \underline{0.324} & \textbf{0.272} & \textbf{0.320} & 0.284 & 0.326 & \underline{0.274} & 0.325 & 0.277 & \underline{0.324} & \underline{0.273} & \underline{0.324} & 0.277 & \underline{0.323} & 0.277 & \underline{0.324} \\
\midrule
\multirow{5}{*}{\rotatebox[origin=c]{90}{ETTh1}}
& 96 & \underline{0.370} & \textbf{0.392} & 0.375 & \textbf{0.392} & \underline{0.371} & \textbf{0.392} & \underline{0.371} & 0.394 & \underline{0.371} & 0.394 & \textbf{0.370} & 0.393 & 0.373 & 0.394 & \underline{0.369} & \textbf{0.392} & \underline{0.370} & 0.393 & \textbf{0.369} & \underline{0.392} \\
& 192 & \underline{0.422} & 0.425 & 0.423 & \textbf{0.423} & \textbf{0.420} & \underline{0.426} & 0.431 & 0.427 & 0.424 & \underline{0.426} & \textbf{0.420} & \textbf{0.423} & 0.424 & 0.425 & 0.421 & 0.424 & \underline{0.419} & \textbf{0.422} & 0.422 & 0.424 \\
& 336 & \textbf{0.443} & \textbf{0.440} & 0.485 & 0.461 & \underline{0.445} & \underline{0.441} & \textbf{0.443} & 0.442 & 0.456 & 0.449 & 0.446 & 0.442 & 0.449 & 0.447 & 0.446 & \underline{0.441} & \underline{0.443} & 0.443 & 0.456 & 0.454 \\
& 720 & \textbf{0.438} & \underline{0.453} & \underline{0.439} & \underline{0.453} & 0.450 & 0.459 & \underline{0.439} & 0.454 & 0.451 & 0.462 & 0.447 & 0.458 & 0.442 & 0.457 & \underline{0.436} & \underline{0.452} & 0.470 & 0.477 & 0.472 & 0.480 \\
\cmidrule(lr){2-22}
& \emph{Avg.} & \textbf{0.418} & \textbf{0.428} & 0.431 & 0.432 & \underline{0.422} & \underline{0.430} & \underline{0.421} & 0.429 & 0.426 & 0.433 & \underline{0.421} & 0.429 & 0.422 & 0.431 & \underline{0.418} & \underline{0.427} & 0.425 & 0.434 & 0.430 & 0.438 \\
\midrule
\multirow{5}{*}{\rotatebox[origin=c]{90}{ETTh2}}
& 96 & \textbf{0.282} & \textbf{0.336} & \underline{0.283} & \textbf{0.336} & 0.286 & 0.339 & \underline{0.283} & 0.337 & 0.284 & 0.338 & 0.286 & 0.339 & 0.287 & 0.340 & 0.287 & 0.342 & 0.286 & 0.338 & 0.291 & 0.343 \\
& 192 & \textbf{0.362} & \textbf{0.388} & \underline{0.367} & 0.392 & 0.373 & 0.394 & 0.368 & 0.391 & 0.370 & 0.393 & 0.373 & 0.395 & 0.370 & 0.393 & 0.371 & 0.393 & 0.376 & 0.395 & 0.378 & 0.396 \\
& 336 & \textbf{0.401} & \textbf{0.421} & 0.424 & 0.431 & 0.427 & 0.429 & \underline{0.414} & 0.424 & 0.426 & 0.432 & 0.422 & 0.427 & \underline{0.406} & \underline{0.420} & 0.425 & 0.430 & 0.424 & 0.428 & 0.434 & 0.432 \\
& 720 & \textbf{0.418} & \textbf{0.439} & 0.439 & 0.451 & 0.439 & 0.449 & 0.425 & 0.442 & \underline{0.419} & \underline{0.438} & 0.435 & 0.445 & 0.434 & 0.447 & 0.435 & 0.446 & 0.429 & 0.443 & 0.435 & 0.452 \\
\cmidrule(lr){2-22}
& \emph{Avg.} & \textbf{0.366} & \textbf{0.396} & 0.378 & 0.403 & 0.381 & 0.403 & \underline{0.373} & 0.399 & \underline{0.375} & 0.400 & 0.379 & 0.402 & \underline{0.374} & \underline{0.400} & 0.380 & 0.403 & 0.379 & 0.401 & 0.384 & 0.405 \\
\midrule
\multirow{5}{*}{\rotatebox[origin=c]{90}{Weather}}
& 96 & \textbf{0.151} & \textbf{0.198} & 0.156 & 0.202 & 0.161 & 0.206 & \underline{0.154} & 0.200 & 0.155 & 0.201 & 0.155 & 0.202 & 0.155 & 0.201 & 0.157 & 0.204 & \underline{0.154} & \underline{0.200} & \underline{0.153} & \underline{0.201} \\
& 192 & \textbf{0.202} & \underline{0.247} & 0.205 & 0.249 & 0.207 & \underline{0.249} & 0.205 & \underline{0.249} & 0.205 & \underline{0.249} & \underline{0.204} & \underline{0.249} & \underline{0.204} & 0.248 & 0.205 & \underline{0.249} & \underline{0.202} & \underline{0.246} & 0.203 & \underline{0.248} \\
& 336 & \textbf{0.260} & \underline{0.292} & 0.265 & \underline{0.291} & 0.265 & \underline{0.292} & \underline{0.262} & \underline{0.292} & 0.261 & \underline{0.290} & 0.264 & \underline{0.291} & 0.268 & 0.293 & \underline{0.262} & \underline{0.289} & \underline{0.260} & \underline{0.290} & 0.264 & 0.293 \\
& 720 & \textbf{0.343} & \underline{0.346} & 0.349 & \underline{0.346} & 0.347 & 0.347 & 0.352 & 0.347 & 0.347 & \underline{0.345} & \underline{0.346} & \underline{0.345} & 0.351 & 0.349 & \underline{0.344} & \underline{0.343} & \underline{0.344} & \underline{0.343} & 0.346 & 0.349 \\
\cmidrule(lr){2-22}
& \emph{Avg.} & \textbf{0.239} & \underline{0.271} & 0.244 & 0.272 & 0.245 & 0.274 & \underline{0.243} & 0.272 & \underline{0.242} & \underline{0.271} & \underline{0.242} & 0.272 & 0.245 & 0.273 & \underline{0.242} & \underline{0.271} & \underline{0.240} & \textbf{0.270} & \underline{0.242} & 0.273 \\
\midrule
\multirow{5}{*}{\rotatebox[origin=c]{90}{Electricity}}
& 96 & \textbf{0.133} & \textbf{0.230} & \underline{0.135} & \textbf{0.230} & 0.149 & 0.240 & 0.140 & 0.237 & \underline{0.135} & 0.231 & 0.148 & 0.240 & 0.137 & 0.232 & 0.171 & 0.260 & 0.138 & 0.235 & \underline{0.140} & \underline{0.235} \\
& 192 & \textbf{0.151} & \textbf{0.245} & \underline{0.154} & \underline{0.247} & 0.163 & 0.253 & 0.159 & 0.253 & 0.156 & 0.249 & \underline{0.150} & \underline{0.244} & \underline{0.154} & \underline{0.247} & 0.170 & 0.261 & 0.153 & 0.249 & 0.163 & 0.255 \\
& 336 & \textbf{0.160} & \textbf{0.257} & \underline{0.162} & \underline{0.259} & 0.181 & 0.271 & 0.175 & 0.275 & 0.165 & 0.263 & 0.163 & 0.260 & \underline{0.163} & 0.260 & 0.197 & 0.286 & 0.165 & 0.265 & 0.170 & 0.268 \\
& 720 & \textbf{0.179} & \textbf{0.279} & \underline{0.183} & \underline{0.281} & 0.217 & 0.306 & 0.235 & 0.321 & 0.193 & 0.286 & 0.182 & 0.282 & 0.184 & 0.281 & 0.198 & 0.292 & 0.198 & 0.291 & 0.196 & 0.291 \\
\cmidrule(lr){2-22}
& \emph{Avg.} & \textbf{0.156} & \textbf{0.253} & \underline{0.159} & \underline{0.254} & 0.178 & 0.268 & 0.177 & 0.272 & \underline{0.162} & 0.257 & \underline{0.161} & 0.257 & \underline{0.160} & \underline{0.255} & 0.184 & 0.275 & 0.164 & 0.260 & 0.167 & 0.262 \\
\midrule
\multirow{5}{*}{\rotatebox[origin=c]{90}{Traffic}}
& 96 & \textbf{0.370} & \textbf{0.247} & \underline{0.372} & 0.250 & 0.412 & 0.261 & 0.379 & 0.251 & 0.380 & 0.252 & \underline{0.372} & \underline{0.249} & 0.374 & \underline{0.250} & 0.485 & 0.325 & 0.381 & 0.252 & 0.381 & 0.253 \\
& 192 & \textbf{0.392} & \textbf{0.258} & \underline{0.396} & 0.262 & 0.425 & 0.267 & 0.401 & \underline{0.214} & 0.403 & 0.263 & 0.395 & 0.261 & \underline{0.396} & 0.262 & 0.400 & 0.263 & 0.405 & 0.263 & 0.401 & 0.263 \\
& 336 & \textbf{0.412} & \textbf{0.269} & \underline{0.413} & \underline{0.271} & 0.440 & 0.274 & 0.416 & 0.272 & 0.416 & \underline{0.271} & 0.414 & 0.273 & \underline{0.414} & \underline{0.272} & 0.417 & \underline{0.271} & 0.416 & \underline{0.271} & 0.417 & \underline{0.271} \\
& 720 & \textbf{0.444} & \underline{0.289} & \underline{0.445} & 0.290 & 0.472 & 0.293 & 0.448 & 0.290 & 0.446 & 0.292 & 0.447 & 0.293 & 0.449 & 0.293 & 0.452 & \underline{0.291} & 0.450 & \underline{0.289} & 0.453 & 0.293 \\
\cmidrule(lr){2-22}
& \emph{Avg.} & \textbf{0.405} & \textbf{0.266} & \underline{0.407} & 0.268 & 0.437 & 0.274 & 0.411 & \underline{0.257} & 0.411 & 0.270 & \underline{0.407} & 0.269 & 0.408 & 0.269 & 0.439 & 0.288 & 0.413 & 0.269 & 0.413 & 0.270 \\
\midrule
\multirow{5}{*}{\rotatebox[origin=c]{90}{Solar-Energy}}
& 96 & \textbf{0.189} & \textbf{0.227} & 0.195 & \underline{0.228} & 0.206 & 0.246 & 0.197 & \underline{0.223} & \underline{0.192} & \underline{0.220} & 0.205 & 0.234 & 0.201 & 0.228 & 0.199 & 0.237 & 0.199 & 0.229 & 0.198 & 0.231 \\
& 192 & \textbf{0.213} & \textbf{0.249} & 0.239 & 0.259 & 0.238 & 0.267 & \underline{0.227} & \underline{0.254} & \underline{0.226} & \underline{0.252} & 0.234 & 0.260 & \underline{0.225} & \underline{0.254} & 0.232 & 0.260 & 0.235 & \underline{0.253} & \underline{0.225} & \underline{0.255} \\
& 336 & \textbf{0.234} & \textbf{0.265} & 0.241 & 0.268 & 0.256 & 0.279 & \underline{0.238} & \underline{0.266} & \underline{0.235} & \underline{0.266} & 0.239 & 0.269 & \underline{0.238} & 0.267 & 0.240 & 0.271 & 0.247 & 0.268 & 0.245 & 0.272 \\
& 720 & \textbf{0.239} & \textbf{0.270} & 0.245 & 0.273 & 0.261 & 0.283 & \underline{0.241} & \textbf{0.270} & \underline{0.240} & \underline{0.271} & 0.247 & 0.275 & 0.243 & 0.276 & 0.248 & 0.275 & 0.244 & \underline{0.271} & 0.246 & 0.273 \\
\cmidrule(lr){2-22}
& \emph{Avg.} & \textbf{0.219} & \textbf{0.253} & 0.230 & 0.257 & 0.240 & 0.269 & \underline{0.226} & \underline{0.253} & \underline{0.223} & \underline{0.252} & 0.231 & 0.260 & \underline{0.227} & \underline{0.256} & 0.230 & 0.261 & 0.231 & \underline{0.255} & \underline{0.229} & 0.258 \\
\bottomrule
\end{tabular}
}
\caption{Full results of ablation study. Ablation analysis showing averaged ETT results and other datasets. results. \textit{Avg} are averaged across three different forecasting horizons: $T \in \{96, 192, 336, 720\}$. The best results are highlighted in \textbf{bold}, while the second-best results are \underline{underlined}.}
\label{tab::app_full_ablation_result}
\end{table*}


\end{document}